\documentclass[twoside,english,3p]{elsarticle}
\usepackage[T1]{fontenc}
\usepackage{geometry}
\usepackage{amssymb}
\usepackage{amsmath}

\usepackage{amsthm}
\usepackage{stmaryrd}
\usepackage{graphicx}
\usepackage{booktabs}
\usepackage{multirow}
\usepackage{makecell}
\usepackage{xurl}
\usepackage{esint}
\usepackage{algorithm}
\usepackage{algpseudocode}
\usepackage{rotating}
\usepackage{adjustbox}
\usepackage{nomencl}
\usepackage{multicol}
\makenomenclature
\makeatletter
\theoremstyle{plain}

\theoremstyle{boldremark} 

\ifx\proof\undefined

\providecommand{\proofname}{Proof}
\fi

\renewenvironment{proof}[1][\proofname]{%
	\par\pushQED{\qed}\normalfont%
	\topsep6\p@\@plus6\p@\relax
	\trivlist\item[\hskip\labelsep\bfseries#1\@addpunct{.}]%
	\ignorespaces
}{%
	\popQED\endtrivlist\@endpefalse
}

\journal{Elsevier}

\usepackage{hyperref}
\hypersetup{colorlinks = true, allcolors = blue}

\usepackage[nameinlink]{cleveref}

\crefname{figure}{Fig.}{Figs.}
\crefformat{equation}{Eq.~#2(#1)#3}
\crefformat{section}{Section~#2#1#3}
\AtBeginDocument{%
	\let\citet\cite
}

\usepackage[labelfont=bf]{caption}
\@ifundefined{showcaptionsetup}{}{%
	\PassOptionsToPackage{caption=false}{subfig}}
\usepackage{subfig}
\makeatother

\usepackage{babel}
\providecommand{\remarkname}{Remark}
\providecommand{\theoremname}{Theorem}

\begin{document}
	
	\begin{frontmatter}{}
		
		\title{Physics-informed Machine Learning for Static Friction Modeling in Robotic Manipulators Based on Kolmogorov-Arnold Networks}

\author[rvt,rvt2]{Yizheng Wang}

\ead{wang-yz19@tsinghua.org.cn}

\author[rvt2]{Timon Rabczuk}

\ead{timon.rabczuk@uni-weimar.de}

\author[rvt]{Yinghua Liu\corref{cor1}}

\ead{yhliu@mail.tsinghua.edu.cn}
\cortext[cor1]{Corresponding author}

\address[rvt]{Department of Engineering Mechanics, Tsinghua University, Beijing 100084, China}

\address[rvt2]{Institute of Structural Mechanics, Bauhaus-Universit\"{a}t Weimar, Marienstr. 15, D-99423 Weimar, Germany}

\begin{abstract}
	Friction modeling plays a crucial role in achieving high-precision motion control in robotic operating systems. Traditional static friction models (such as the Stribeck model) are widely used due to their simple forms; however, they typically require predefined functional assumptions, which poses significant challenges when dealing with unknown functional structures. To address this issue, this paper proposes a physics-inspired machine learning approach based on the Kolmogorov–Arnold Network (KAN) for static friction modeling of robotic joints. The method integrates spline activation functions with a symbolic regression mechanism, enabling model simplification and physical expression extraction through pruning and attribute scoring, while maintaining both high prediction accuracy and interpretability. We first validate the method’s capability to accurately identify key parameters under known functional models, and further demonstrate its robustness and generalization ability under conditions with unknown functional structures and noisy data. Experiments conducted on both synthetic data and real friction data collected from a six-degree-of-freedom industrial manipulator show that the proposed method achieves a coefficient of determination greater than 0.95 across various tasks and successfully extracts concise and physically meaningful friction expressions. This study provides a new perspective for interpretable and data-driven robotic friction modeling with promising engineering applicability.
\end{abstract}

		\printnomenclature

		
	\end{frontmatter}{}

\section{Introduction}

In robotic operating systems, friction plays a crucial role in determining motion control accuracy, particularly in high-precision, low-velocity, and force-controlled tasks, where its influence becomes markedly pronounced.
Traditionally, friction models are broadly classified into static and dynamic categories.
Static models, such as the Coulomb and Stribeck models \cite{lu2006stribeck}, describe friction as a function of velocity and normal load, without accounting for any historical dependence. Although these models are simple and easy to implement, they fail to capture key nonlinear phenomena commonly observed in real frictional behavior—such as stick–slip motion, hysteresis, and frictional lag \cite{armstrong1992frictional}.
In contrast, dynamic friction models—including the Dahl, LuGre, and Generalized Maxwell–Slip (GMS) models \cite{lampaert2003generalized,piatkowski2014dahl}—introduce internal state variables to more accurately represent the time-dependent evolution of friction.
However, despite their improved realism, these models are often structurally complex and highly parameterized, making model identification and real-time computation challenging. Such limitations restrict their practical applicability in engineering implementations.

This study focuses on static friction modeling, aiming to leverage neural network techniques to automatically regress physically interpretable friction laws from experimental data.
In traditional approaches, the identification of static friction models typically requires the functional form of the friction law to be predefined, which relies heavily on expert knowledge and experience. For different types of robotic manipulators, practitioners must often select or tailor the most appropriate functional form, demanding a high level of domain expertise and extensive empirical insight.
Over the past decade, artificial intelligence (AI) has profoundly transformed numerous scientific and engineering domains, including computer vision \cite{ALEXNET}, speech recognition \cite{speech_recognition}, natural language processing \cite{brown2020language}, strategic gaming \cite{alphago,star_game}, and drug discovery \cite{alphafold}. Indeed, deep learning has become ubiquitous, driving innovation across a wide spectrum of disciplines.
A prominent paradigm involves training deep neural networks on experimental measurements or high-fidelity numerical data \cite{li2019predicting} to construct surrogate models \cite{ill_gradient}. These models can learn to represent complex, nonlinear relationships directly from data through their trainable parameters.
However, the knowledge embedded in such networks is often opaque and difficult to interpret, as the learned representations are encoded within large sets of black-box parameters, limiting their physical transparency and interpretability.

Recently, Kolmogorov–Arnold Networks (KANs) \cite{liu2024kan} have emerged as a novel neural network architecture that employs B-spline activations, naturally supporting symbolic regression and network pruning, thereby offering both strong interpretability and high expressive capacity.
In this work, we incorporate KAN into the static friction identification of robotic manipulators, aiming to discover physically consistent and compact friction laws directly from data.
We demonstrate that KAN can effectively learn concise and physically interpretable friction representations, even when the underlying analytical form is completely unknown.
The proposed approach is validated through both synthetic datasets and experimental measurements collected from industrial robotic manipulators.
Our experiments encompass a broad range of tasks, including parameter identification for known functional forms, data-driven modeling under unknown structures, robustness analysis under noisy conditions, and generalization from single-axis to multi-axis scenarios.
The results confirm that the KAN-based framework achieves superior performance in terms of accuracy, interpretability, and generalization capability, highlighting its potential for advancing intelligent friction modeling in robotics.

The outline of the paper is as follows. \Cref{sec:introduction_friction_model} introduces classical static friction models in robotic manipulators. \Cref{sec:method} presents the proposed approach based on Kolmogorov–Arnold Networks (KAN) for identifying static friction models. \Cref{sec:results} validates the method using both synthetic data and real friction measurements from a robotic manipulator. Finally, \Cref{sec:conclusion} summarizes the current advantages and limitations of applying KAN to static friction model identification.

\section{Static Friction Models in Robotic Manipulators\label{sec:introduction_friction_model}}

Friction models in robotic manipulators can be broadly divided into two categories: static and dynamic models.  

In general, static friction models are suitable for relatively simple structures and stable motion states. In such models, the friction force typically depends only on the current velocity and normal load, without considering the history of motion. However, in practical scenarios, frictional processes often exhibit history-dependent characteristics, such as pre-sliding displacement, hysteresis loops, and stick–slip motion. In these cases, static models may fail to accurately capture the variations in friction force, potentially leading to significant modeling errors. By contrast, dynamic friction models introduce internal state variables to account for historical information such as motion trajectories and velocity variations. These models are closer to physical reality and are capable of reproducing key behaviors, including the Stribeck effect, stick–slip transitions, hysteresis, and pre-sliding displacement. Nevertheless, the increased structural complexity and large number of parameters make dynamic models more challenging to identify and solve, as they are typically expressed in the form of differential equations.  

\begin{figure}
	\begin{centering}
		\includegraphics[scale=0.45]{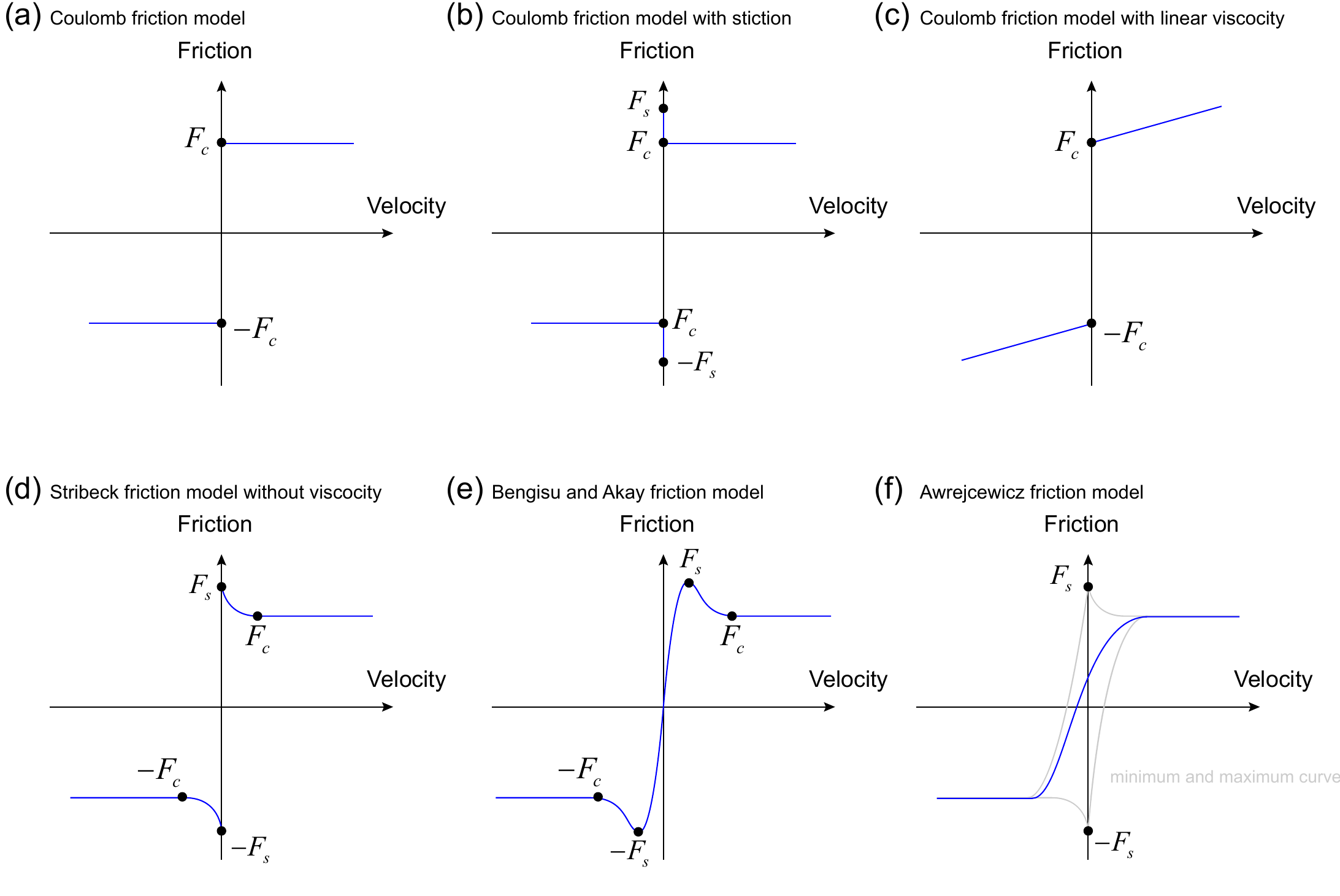}
		\par\end{centering}
	\caption{Classical static friction models, where velocity denotes tangential relative motion: (a) classical Coulomb model \cite{coulomb1781theorie}, (b) Coulomb model with static friction, (c) Coulomb model with viscous term, (d) Stribeck model without viscosity, (e) smoothed Bengisu–Akay model \cite{bengisu1994stability}, (f) Awrejcewicz envelope model, with the gray line showing the model’s envelope range \cite{awrejcewicz2008404}. \label{fig:classic_static_friction_models}}
\end{figure}

The present study focuses on static friction modeling, aiming to infer friction models directly from experimental data. Among static friction models, the most widely adopted is the Stribeck-type formulation \cite{bo1982friction}. As shown in \Cref{fig:classic_static_friction_models}, classical static models have evolved over time, ranging from the simple Coulomb model \cite{coulomb1781theorie} to more sophisticated representations, including the Bengisu–Akay \cite{bengisu1994stability} and Awrejcewicz models \cite{awrejcewicz2008404}. For a comprehensive survey of classical models, see \cite{marques2016survey}. Despite the diversity of models, the Stribeck formulation remains the most commonly used, as it captures the transition between static and kinetic friction:  
\begin{equation}
	F(v) = \left[F_{c} + (F_{s}-F_{c}) \exp\!\left(-\frac{|v|}{v_{s}}\right)\right]\mathrm{sign}(v) + F_{v}v,
	\label{eq:static_friction_model}
\end{equation}
where $F$ denotes the friction force, $v$ is the velocity, $F_{c}$ is the Coulomb friction coefficient, $F_{s}$ is the static friction coefficient, and $F_{v}$ is the viscous friction coefficient corresponding to fluid-type lubrication. The parameter $v_{s}$ is the Stribeck velocity.  

\begin{figure}
	\begin{centering}
		\includegraphics[scale=0.45]{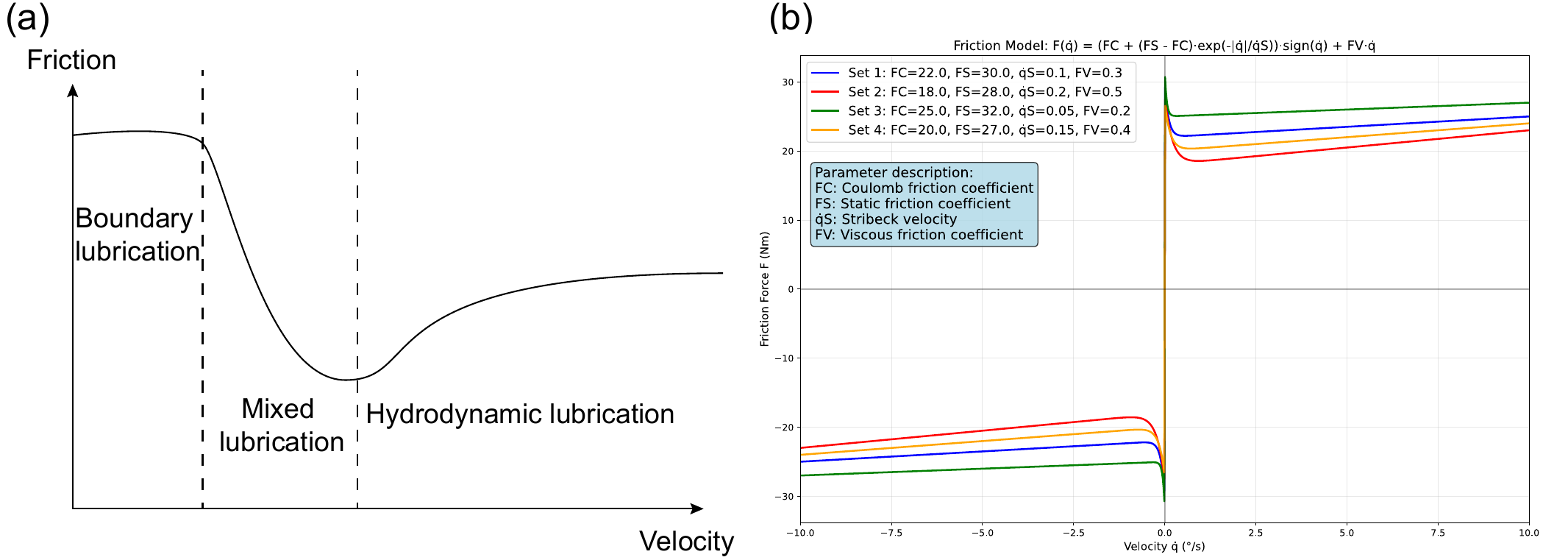}
		\par\end{centering}
	\caption{The Stribeck effect: (a) schematic illustration of the Stribeck curve, showing boundary, mixed, and hydrodynamic lubrication regimes \cite{he2017experimental}; (b) velocity–friction predictions of the static Stribeck-type model. \label{fig:Stribeck}}
\end{figure}

The Stribeck effect is a well-known phenomenon in tribology \cite{lu2006stribeck}, which allows for a continuous transition between static and dynamic friction, in contrast to the abrupt change in Coulomb friction (\Cref{fig:classic_static_friction_models}b). The Stribeck curve is typically divided into three lubrication regimes—boundary, mixed, and hydrodynamic lubrication—as illustrated in \Cref{fig:Stribeck}a. In the boundary regime, surfaces are in direct contact with minimal lubrication, leading to high friction and wear. In the mixed regime, partial fluid films form, reducing friction significantly. In the hydrodynamic regime, a complete lubrication film separates the surfaces, minimizing both friction and wear.  

\Cref{fig:Stribeck}b shows the velocity–friction relation predicted by \Cref{eq:static_friction_model}, which accurately reproduces the mixed and hydrodynamic lubrication regimes. However, the fitting performance in the low-velocity region (especially near $v=0$, corresponding to the boundary lubrication regime) is often unsatisfactory.  

To better capture the boundary regime and the overall friction process, dynamic models are often employed. From a physical perspective, static models fail to incorporate important phenomena such as self-excited vibrations and frictional lag. The term “frictional lag” refers to the time delay between changes in velocity and the corresponding changes in friction force \cite{hess1990friction}, indicating that friction inherently depends on history. This limitation highlights the necessity of dynamic friction models that explicitly incorporate memory effects.  

\section{Method}\label{sec:method}

\subsection{KAN Network with Multiplication}

We first introduce the structure of the Kolmogorov–Arnold Network (KAN) \cite{liu2024kan,liu2024kan2}. In KAN, suppose the number of input neurons in a given layer is $l_{i}$ and the number of output neurons is $l_{o}$. The activation function in this layer is denoted as $\phi_{ij}$, where $i\in\{1,2,\cdots,l_{o}\}$ and $j\in\{1,2,\cdots,l_{i}\}$. Each element of the spline-based activation function $\phi_{ij}$ is determined by the grid number $G$ and the order $r$ of the B-spline, and is expressed as  
\begin{equation}
	\phi_{ij}(\boldsymbol{X})=\left[\begin{array}{cccc}
		\sum_{m=1}^{G_{1}+r_{1}}c_{m}^{(1,1)}B_{m}(x_{1}) & \sum_{m=1}^{G_{2}+r_{2}}c_{m}^{(1,2)}B_{m}(x_{2}) & \cdots & \sum_{m=1}^{G_{l_{i}}+r_{l_{i}}}c_{m}^{(1,l_{i})}B_{m}(x_{l_{i}})\\
		\sum_{m=1}^{G_{1}+r_{1}}c_{m}^{(2,1)}B_{m}(x_{1}) & \sum_{m=1}^{G_{2}+r_{2}}c_{m}^{(2,2)}B_{m}(x_{2}) & \cdots & \sum_{m=1}^{G_{l_{i}}+r_{l_{i}}}c_{m}^{(2,l_{i})}B_{m}(x_{l_{i}})\\
		\vdots & \vdots & \ddots & \vdots\\
		\sum_{m=1}^{G_{1}+r_{1}}c_{m}^{(l_{o},1)}B_{m}(x_{1}) & \sum_{m=1}^{G_{2}+r_{2}}c_{m}^{(l_{o},2)}B_{m}(x_{2}) & \cdots & \sum_{m=1}^{G_{l_{i}}+r_{l_{i}}}c_{m}^{(l_{o},l_{i})}B_{m}(x_{l_{i}})
	\end{array}\right],
\end{equation}
where $G_{j}$ denotes the number of grids in the $j$-th input direction, $r_{j}$ denotes the B-spline order in that direction, and $j\in\{1,2,\cdots,l_{i}\}$. The coefficients $c_{m}^{(i,j)}$ correspond to the B-spline basis functions $B_{m}$, with a total number of $G_{j}+r_{j}$ for each input direction. It is important to note that both the grid partitioning and the order of the B-splines can be chosen independently for each input direction.  

To enhance the fitting capability of the activation functions, we introduce an additional scaling matrix $S_{ij}$, which has the same dimensions as $\phi_{ij}$:  
\begin{equation}
	S_{ij}=\left[\begin{array}{cccc}
		s_{11} & s_{12} & \cdots & s_{1l_{i}}\\
		s_{21} & s_{22} & \cdots & s_{2l_{i}}\\
		\vdots & \vdots & \ddots & \vdots\\
		s_{l_{o}1} & s_{l_{o}2} & \cdots & s_{l_{o}l_{i}}
	\end{array}\right].
\end{equation}
The role of $\boldsymbol{S}$ is to adjust the amplitude of the activation function $\boldsymbol{\phi}$, such that $\boldsymbol{\phi}=\boldsymbol{\phi}\odot\boldsymbol{S}$, where $\odot$ denotes the element-wise product. The final output of the layer is expressed as  
\begin{equation}
	\boldsymbol{Y}=\tanh\left\{\sum_{\text{column}}[\boldsymbol{\phi}(\boldsymbol{X})\odot\boldsymbol{S}]+\boldsymbol{W}\cdot\sigma(\boldsymbol{X})\right\},
\end{equation}
where $\boldsymbol{W}$ is a linear weight matrix, and $\sigma$ denotes a nonlinear activation function. The inclusion of $\sigma$ ensures smoothness of the fitted function, since relying solely on B-splines may lead to rough approximations. Both $\boldsymbol{S}$ and $\boldsymbol{W}$ act as scaling factors, analogous to normalization techniques commonly used in machine learning. The residual term $\boldsymbol{W}\cdot\sigma(\boldsymbol{X})$ is inspired by the residual learning framework introduced in ResNet \cite{he2016deep}.  

If we assume that the grid partitioning and spline orders are identical across all input directions, then the number of coefficients $c_{m}^{(i,j)}$ is consistently $G+r$. The trainable parameters in KAN are summarized in \Cref{tab:trainable_para_KAN}, and further details can be found in the original KAN paper \cite{liu2024kan}.  

\begin{table}
	\caption{Trainable parameters in KAN \label{tab:trainable_para_KAN}}
	\centering{}%
    \begin{adjustbox}{max width=\textwidth}
	\begin{tabular}{|c|c|c|c|}
		\hline 
		Parameter type & Variable & Count & Description \tabularnewline
		\hline 
		$c_{m}^{(i,j)}$ & spline\_weight & $l_{o}\times l_{i}\times(G+r)$ & B-spline coefficients in activation functions $\boldsymbol{\phi}$ \tabularnewline
		\hline 
		$W_{ij}$ & base\_weight & $l_{i}\times l_{o}$ & Linear transformation applied after nonlinear activation $\sigma$ \tabularnewline
		\hline 
		$S_{ij}$ & spline\_scaler & $l_{i}\times l_{o}$ & Scaling factor for adjusting the amplitude of activation functions \tabularnewline
		\hline 
	\end{tabular}
    \end{adjustbox}
\end{table}

\subsection{Inverse Identification of Static Friction in Robotic Manipulators with KAN}
\label{subsec:KAN_inverse}

\begin{figure}
	\begin{centering}
		\includegraphics[scale=0.45]{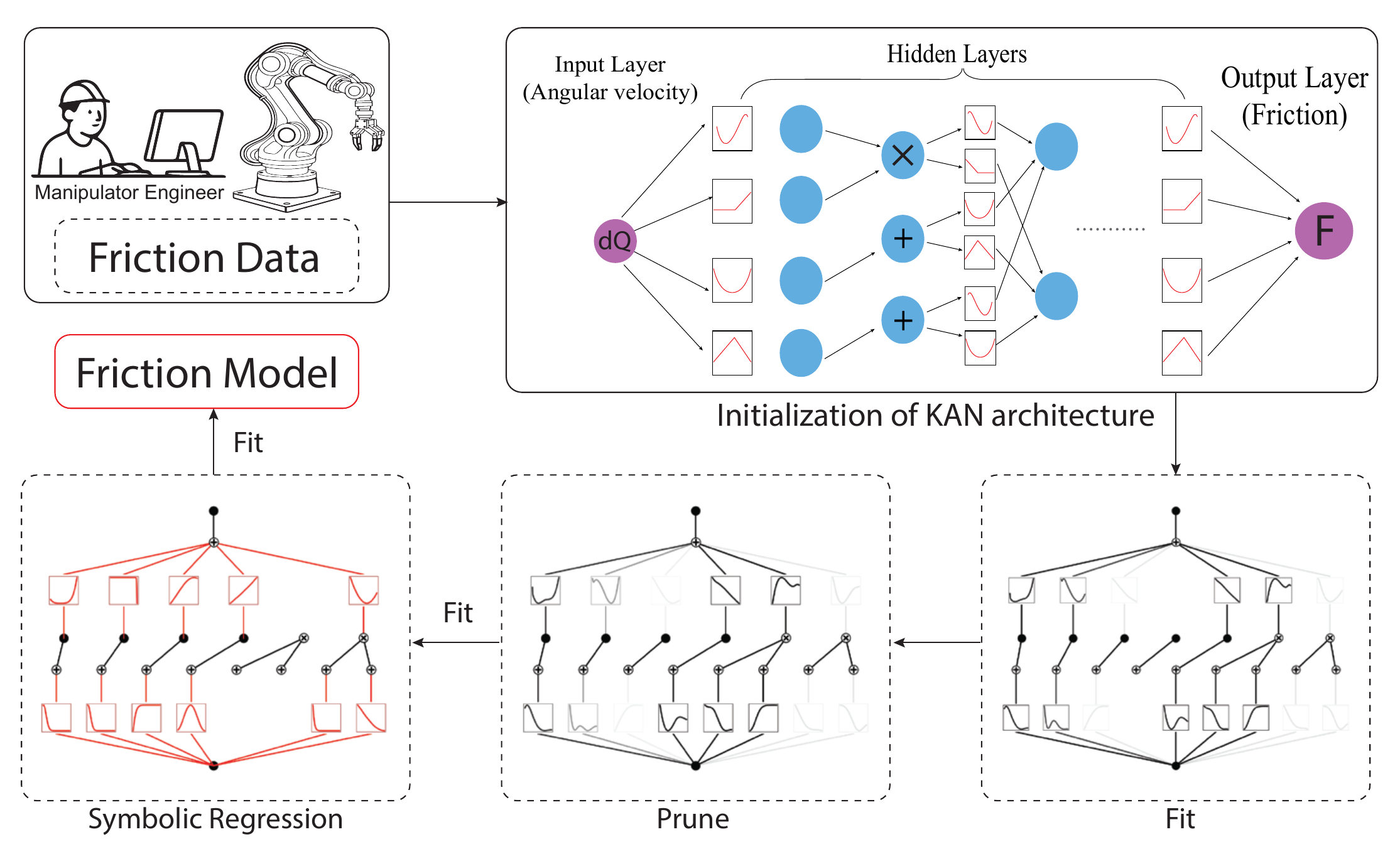}
		\par\end{centering}
	\caption{Schematic of KAN-based friction model identification, consisting of six steps: (1) collection of friction data by robotic engineers, (2) initialization of the KAN network, (3) model fitting with KAN, (4) pruning followed by refitting, (5) symbolic regression and subsequent refitting, and (6) derivation of the final friction model. \label{fig:KAN_methodology_friction}}
\end{figure}

To illustrate the entire KAN-based static friction identification workflow, the procedure can be summarized in six key steps (see \Cref{fig:KAN_methodology_friction}):

\begin{enumerate}
    \item \textbf{Data acquisition} – frictional force data are collected from the robotic manipulator by engineers;
    \item \textbf{Network initialization} – the KAN architecture is initialized with selected hyperparameters;
    \item \textbf{Initial training} – the KAN model is trained to fit the collected friction data;
    \item \textbf{Pruning} – redundant nodes and connections are removed based on importance scoring, followed by fine-tuning;
    \item \textbf{Symbolic regression} – symbolic expressions are extracted from the pruned network and further refined through retraining;
    \item \textbf{Model formulation} – the final, physically interpretable friction law is obtained.
\end{enumerate}

The pruning process in KAN serves as a network simplification mechanism grounded in importance scoring. 
First, the attribute score of each node and edge is computed to quantify their contribution to the overall network output.
During node pruning, neurons with attribute scores below a predefined threshold are eliminated. 
Special care is taken in handling additive and multiplicative nodes; for multiplicative nodes in particular, their importance scores must be appropriately propagated to the corresponding child nodes to maintain consistent contribution tracking.
Edge pruning is subsequently performed by updating a mask matrix, setting insignificant connections to zero while retaining those that contribute meaningfully to the output. 
In addition, input feature pruning specifically targets the input layer, removing variables that exert negligible influence on the model predictions.

The entire pruning workflow follows a bottom-up strategy: node pruning is first performed to simplify the network topology, after which attribute scores are recomputed for subsequent edge pruning. 
Through this hierarchical process, a more compact and computationally efficient KAN model is obtained while preserving the essential predictive capability of the original architecture.

Overall, this workflow highlights how KAN enables a data-driven yet interpretable framework for friction modeling, effectively bridging the gap between machine learning representations and physical understanding.

\section{Results of Static Friction Modeling}\label{sec:results}

\subsection{Prediction of Data-Driven Friction Models under Known Functional Forms\label{subsec:known_function}}

\begin{figure}
	\begin{centering}
		\includegraphics[scale=0.45]{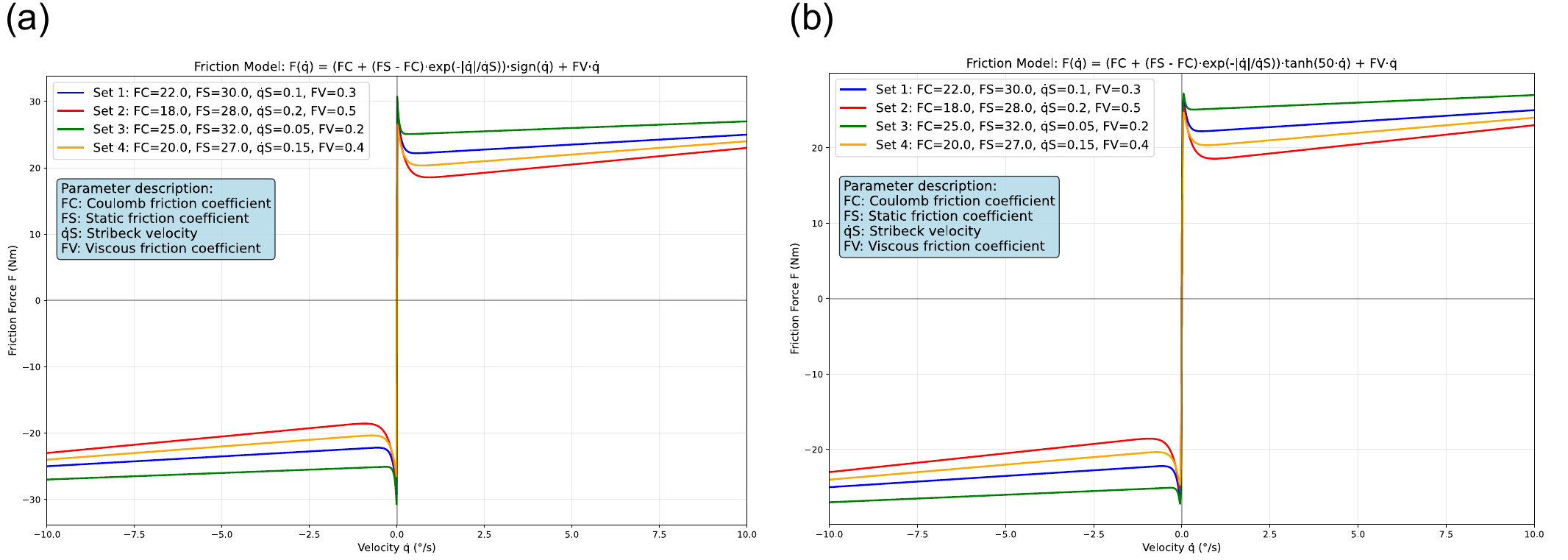}
		\par\end{centering}
	\caption{Comparison between the original $\mathrm{sign}$ and the smoothed $\tanh$ in the Stribeck static friction model. \label{fig:Stribeck_sign_tanh}}
\end{figure}

We first evaluate the proposed framework under the condition of a known functional form, where only specific parameters of the model are to be identified. It is worth noting that in this case, the functional structure of the friction model is assumed to be known, and only a few constant coefficients are estimated. We adopt the classical Stribeck friction model, as shown in \Cref{eq:static_friction_model}. However, due to the discontinuity of the $\mathrm{sign}(v)$ function at $v=0$, numerical instability may arise \cite{marques2016survey}. To address this issue, we introduce a smooth approximation by replacing $\mathrm{sign}(v)$ with $\tanh$, yielding the following modified formulation:  
\begin{equation}
	\begin{aligned}F(v) & =[F_{c}+(F_{s}-F_{c})\exp(-\tfrac{|v|}{v_{s}})]\mathrm{sign}(v)+F_{v}v,\\
		F(v) & =(k_{1}+k_{2}\exp[-|\tfrac{v}{k_{3}}|])\tanh(50v)+k_{4}v,
	\end{aligned}
	\label{eq:stribeck_friction_model}
\end{equation}
where $F$ is the friction force and $v$ is the angular velocity. \Cref{fig:Stribeck_sign_tanh} compares the original $\mathrm{sign}$ formulation and the smoothed $\tanh$ function, showing that the curves coincide perfectly while the $\tanh$ ensures differentiability everywhere. The coefficients are related as $k_{1}=F_{c}$, $k_{2}=F_{s}-F_{c}$, $k_{3}=v_{s}$, and $k_{4}=F_{v}$, which need to be identified from $N$ data samples $\{v^{(i)},F^{(i)}\}_{i=1}^{N}$. For the experimental setup, the parameters are selected as listed in \Cref{tab:parameters_friction_model}.  

\begin{table}
	\caption{Coefficients to be identified in the friction model. \label{tab:parameters_friction_model}}
	\centering{}%
	\begin{tabular}{ccccccc}
		\toprule 
		Axis & 1 & 2 & 3 & 4 & 5 & 6\tabularnewline
		\midrule
		$k_{1}$ & 22 & 23 & 24 & 25 & 26 & 27\tabularnewline
		$k_{2}$ & 8 & 10 & 12 & 14 & 16 & 18\tabularnewline
		$k_{3}$ & 0.10 & 0.13 & 0.16 & 0.19 & 0.22 & 0.25\tabularnewline
		$k_{4}$ & 0.03 & 0.16 & 0.29 & 0.42 & 0.55 & 0.68\tabularnewline
		\bottomrule
	\end{tabular}
\end{table}

We then estimate $\{k_{1},k_{2},k_{3},k_{4}\}$ using a standard regression framework:  
\begin{equation}
	\begin{aligned}
		\{k_{1},k_{2},k_{3},k_{4}\}  &= \underset{\{k_{1},k_{2},k_{3},k_{4}\}}{\arg\min}\, \mathcal{L},\\
		\mathcal{L} & =\frac{1}{N}\sum_{i=1}^{N}\left[(k_{1}+k_{2}\exp[-|\tfrac{v^{(i)}}{k_{3}}|])\tanh(50v^{(i)})+k_{4}v^{(i)}-F^{(i)}\right]^{2}.
	\end{aligned}
\end{equation}

\begin{figure}
	\begin{centering}
		\includegraphics[scale=0.45]{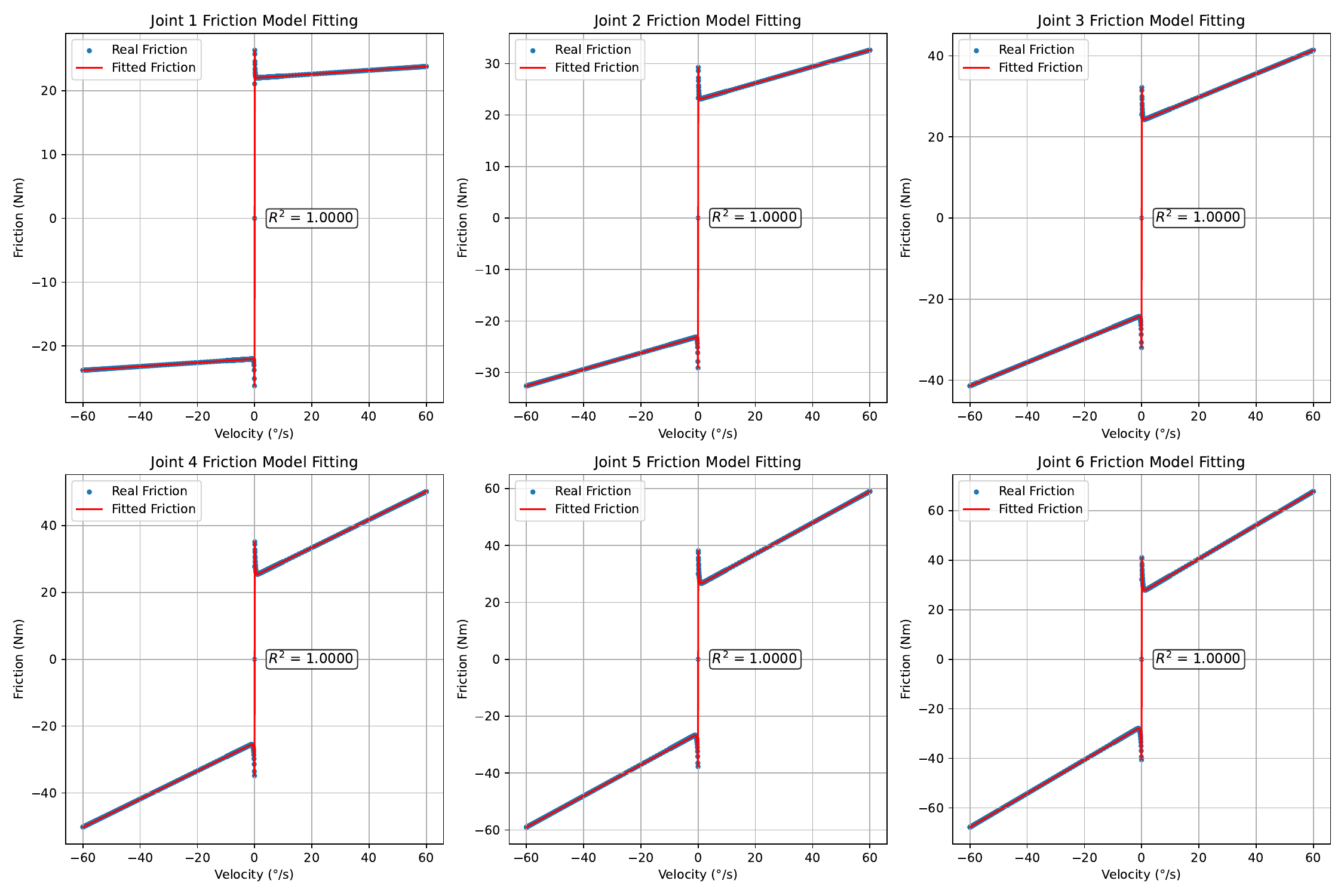}
		\par\end{centering}
	\caption{Predictions of the friction torque model across six axes. The coefficient of determination $R^{2}$ is used as the regression metric. \label{fig:prediction_m_f}}
\end{figure}

The optimization is performed using the Adam optimizer with a learning rate of $0.01$ and a default of 30,000 iterations. The dataset contains $N=1000$ samples. \Cref{fig:prediction_m_f} illustrates the prediction results of the six axes after 30,000 iterations, showing excellent agreement between the model and the original data. The goodness of fit is quantified by the coefficient of determination $R^{2}$, defined as:  
\begin{equation}
	R^{2}=1-\frac{\sum_{i=1}^{N}(y_{i}-\hat{y}_{i})^{2}}{\sum_{i=1}^{N}(y_{i}-\bar{y})^{2}},
\end{equation}
where $y_{i}$ and $\hat{y}_{i}$ are the ground truth and predicted values, respectively, and $\bar{y}$ is their mean.  

To further assess performance, we vary the dataset size from the original $N=1000$ samples. \Cref{tab:performance_different_data} summarizes the relative errors $\mathcal{L}_{rel}$ of parameter estimation after 30,000 iterations, where  
\begin{equation}
	\mathcal{L}_{rel}=\left|\frac{k^{\text{pred}}-k^{\text{true}}}{k^{\text{true}}}\right|,
\end{equation}
with $k^{\text{pred}}$ and $k^{\text{true}}$ denoting predicted and ground-truth parameters, respectively. The results indicate that even with a reduced dataset, the model maintains remarkable accuracy, achieving nearly perfect identification of the parameters $k_{i}$.  

\begin{table}
	\caption{Relative error $\mathcal{L}_{rel}$ of the friction model predictions under different dataset sizes after 30,000 iterations. \label{tab:performance_different_data}}
	\centering{}%
    \begin{adjustbox}{max width=\textwidth}
	\begin{tabular}{ccccccccccccc}
		\toprule 
		\multirow{2}{*}{Axis} & \multicolumn{3}{c}{$k_{1}$} & \multicolumn{3}{c}{$k_{2}$} & \multicolumn{3}{c}{$k_{3}$} & \multicolumn{3}{c}{$k_{4}$}\tabularnewline
		\cmidrule{2-13} 
		& $10\%$ & $30\%$ & $80\%$ & $10\%$ & $30\%$ & $80\%$ & $10\%$ & $30\%$ & $80\%$ & $10\%$ & $30\%$ & $80\%$\tabularnewline
		\midrule
		1 & 7.33e-6 & 2.94e-6 & 6.54e-6 & 4.65e-6 & 3.93e-6 & 8.27e-6 & 9.47e-6 & 2.43e-6 & 1.78e-6 & 3.94e-6 & 2.64e-6 & 7.94e-6\tabularnewline
		2 & 3.43e-6 & 6.56e-6 & 1.24e-6 & 6.54e-6 & 2.56e-6 & 6.73e-6 & 8.24e-6 & 1.26e-6 & 5.70e-6 & 1.27e-6 & 1.06e-6 & 6.70e-6\tabularnewline
		3 & 4.38e-6 & 8.87e-6 & 2.65e-6 & 2.21e-6 & 8.62e-6 & 2.95e-6 & 8.74e-6 & 6.56e-6 & 7.83e-6 & 2.70e-6 & 5.08e-6 & 3.96e-6\tabularnewline
		4 & 8.74e-6 & 1.34e-6 & 7.76e-6 & 8.57e-6 & 2.74e-6 & 5.74e-6 & 1.26e-6 & 4.24e-6 & 7.30e-6 & 1.30e-6 & 1.89e-6 & 1.38e-6\tabularnewline
		5 & 4.28e-6 & 4.67e-6 & 5.98e-6 & 2.96e-6 & 6.62e-6 & 4.22e-6 & 2.65e-6 & 7.44e-6 & 6.92e-6 & 4.34e-6 & 2.25e-6 & 2.80e-6\tabularnewline
		6 & 5.63e-6 & 3.59e-6 & 4.76e-6 & 5.95e-6 & 5.84e-6 & 1.35e-6 & 4.75e-6 & 2.88e-6 & 3.45e-6 & 4.09e-6 & 1.56e-6 & 4.01e-6\tabularnewline
		\bottomrule
	\end{tabular}
    \end{adjustbox}
\end{table}

Next, we examine the robustness of the model by introducing noise, which reflects practical experimental conditions. The noise distribution is Gaussian with standard deviation defined as  
\begin{equation}
	\text{Std}=25\%\,[\max(F)-\min(F)], \label{eq:std_equation}
\end{equation}
i.e., 25\% of the range of friction torque. The noisy dataset is generated as:  
\[
F_{\text{noisy}}=F_{\text{origin}}+N(0,\text{Std}),
\]
where $N(0,\text{Std})$ denotes Gaussian noise with zero mean and standard deviation $\text{Std}$.  

\begin{figure}
	\begin{centering}
		\includegraphics[scale=0.45]{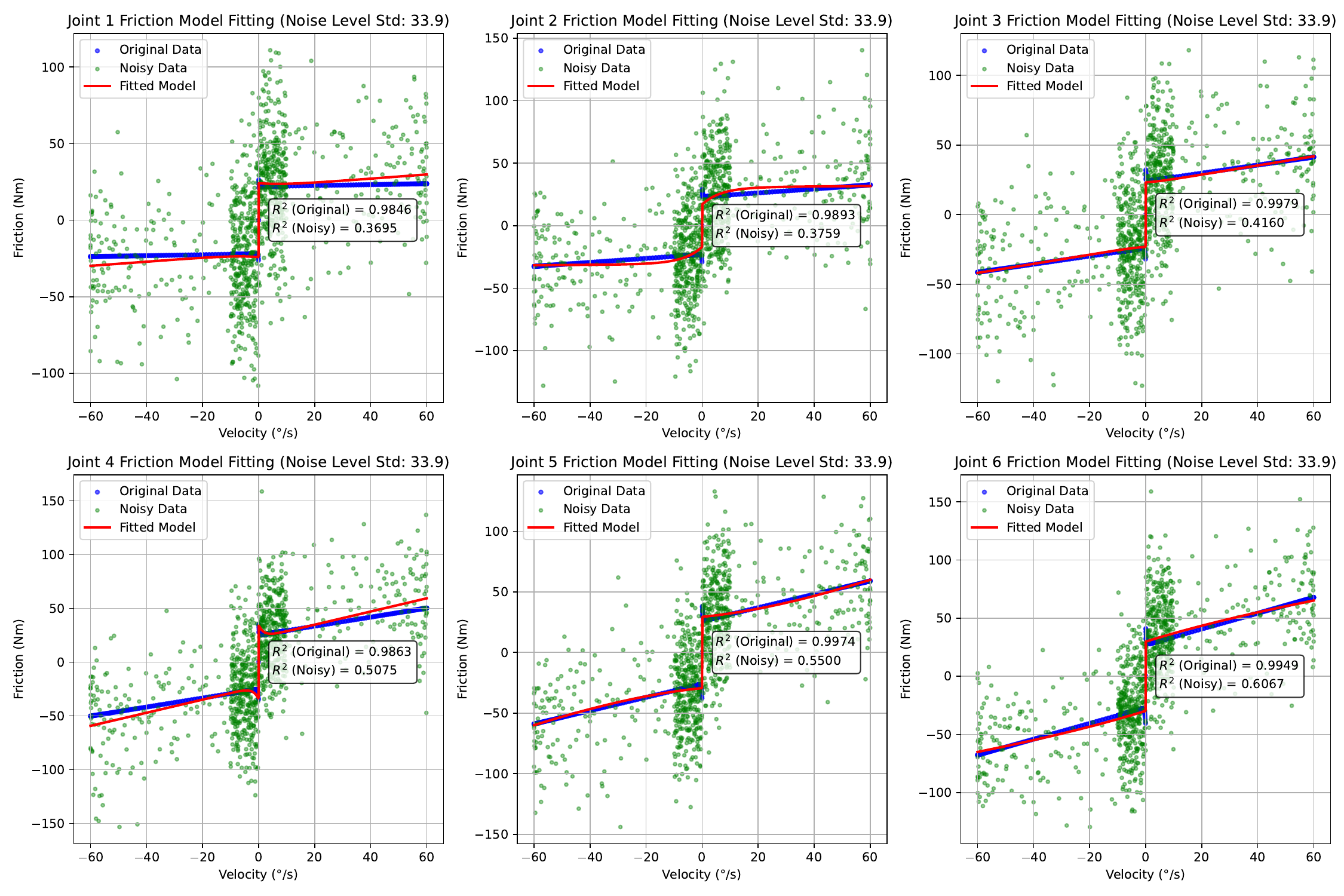}
		\par\end{centering}
	\caption{Prediction of the friction torque model with noisy data. $R^{2}$(Original) denotes the coefficient of determination evaluated on the original clean data after training on noisy data, while $R^{2}$(Noisy) denotes the coefficient of determination evaluated directly on noisy data. \label{fig:prediction_m_f_noisy}}
\end{figure}

\Cref{fig:prediction_m_f_noisy} shows the prediction performance with noisy data. The results demonstrate that the model exhibits strong robustness: even under noise levels as high as 25\% of the torque range, the predictions remain in excellent agreement with the original noise-free data. It should be emphasized that the model was trained entirely on noisy data.  

\subsection{Data-Driven Friction Modeling under Unknown Functional Forms\label{subsec:unknown_function}}

This section demonstrates two types of generalization: (i) trajectory-to-trajectory, where the friction model is learned on one trajectory and then generalized to other trajectories; and (ii) single-axis to multi-axis, where the model is trained on one joint and generalized to multiple joints. Unlike \Cref{subsec:known_function}, we assume here that the functional form of the friction model is entirely unknown.

We first validate the approach on synthetic data, and then apply it to real experimental measurements collected from an HSR-JR607 industrial manipulator.

\subsubsection{Validation on Synthetic Data}

We begin with synthetic data generated from the Stribeck friction equation:
\begin{equation}
	\begin{aligned}
		F(v) &= \big[F_{c}+(F_{s}-F_{c})\exp\!\big(-\tfrac{|v|}{v_{s}}\big)\big]\tanh(50v)+F_{v}v,\\
		F(v) &= \big(k_{1}+k_{2}\exp[-|\tfrac{v}{k_{3}}|]\big)\tanh(50v)+k_{4}v,
	\end{aligned}
	\label{eq:stribeck_friction_model-1}
\end{equation}
where $v$ is the relative sliding velocity, and the physical meanings of $\{k_{1},k_{2},k_{3},k_{4}\}$ follow \Cref{eq:stribeck_friction_model}. We design six parameter sets $\{k_{1},k_{2},k_{3},k_{4}\}$ to represent the friction characteristics of six joints. For each set, we uniformly sample $v\in[-1,1]$ to generate paired data $\{v^{(i)},F^{(i)}\}_{i=1}^{N}$ for supervised learning with $N=1000$.

\begin{figure}
	\begin{centering}
		\includegraphics[scale=0.35]{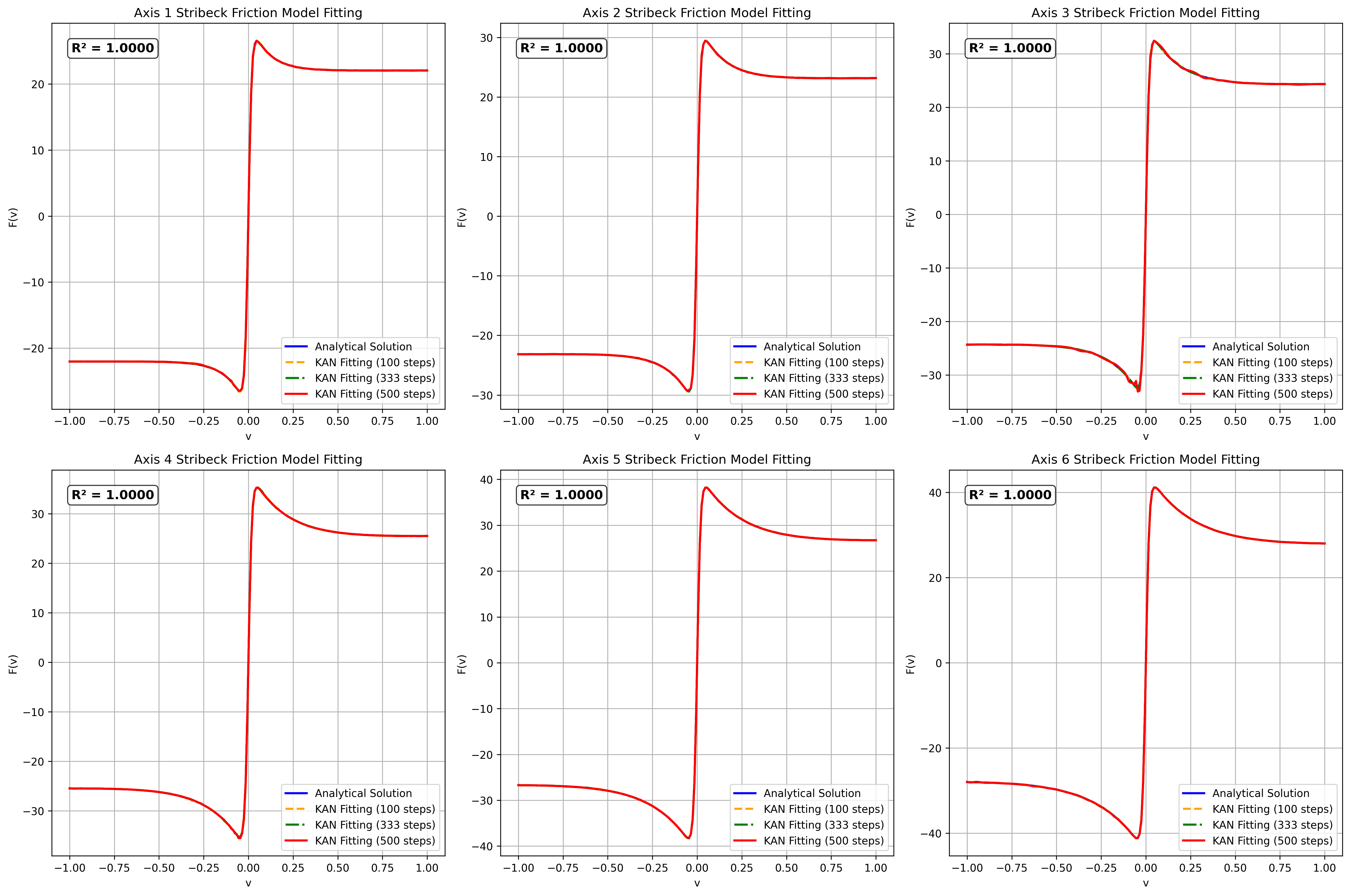}
		\par\end{centering}
	\caption{KAN fitting of a friction model with unknown functional form  of six joints. \label{fig:KAN_fitting_unknown}}
\end{figure}

The model uses a KAN architecture of $[1,5,1]$, with 10 grid points per dimension. Training is performed using the L-BFGS optimizer for a total of 300 steps. To analyze the dynamics of fitting, we record intermediate results at $1/3$, $2/3$, and all training steps, plot the fitted curves against the analytical solution, and report the final coefficient of determination $R^{2}$ as the accuracy metric. As shown in \Cref{fig:KAN_fitting_unknown}, the KAN achieves highly accurate fits to the Stribeck behavior.

\begin{figure}
	\begin{centering}
		\includegraphics[scale=0.45]{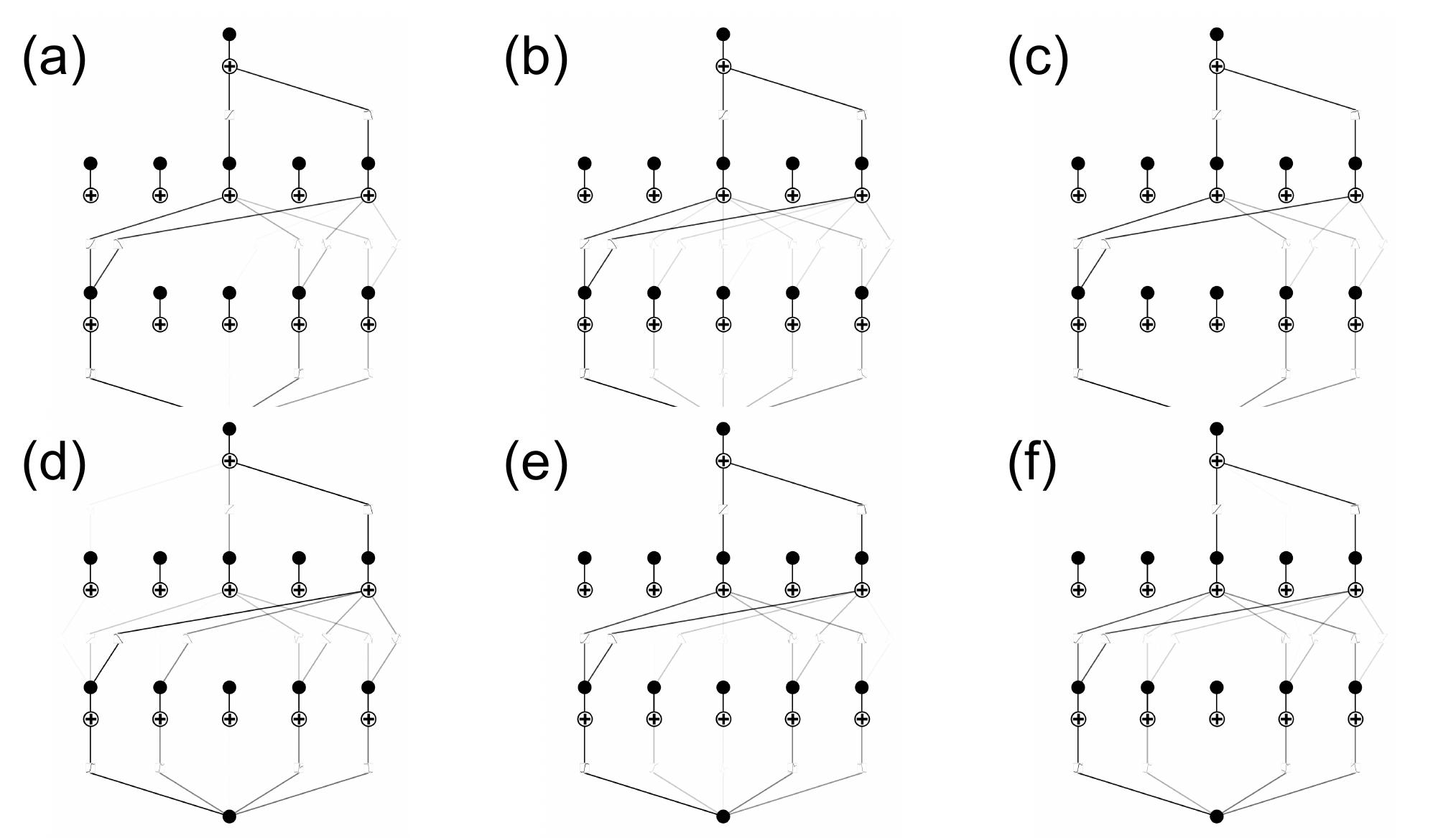}
		\par\end{centering}
	\caption{Activation functions of KAN nodes (B-spline forms) in different 6 joints. \label{fig:activation_function_kan}}
\end{figure}

An important advantage of KAN is its ability to support symbolic representation. \Cref{fig:activation_function_kan} visualizes the B-spline-based activation functions learned by KAN, highlighting their flexibility. However, due to the complexity of B-spline expressions, practical engineering applications often prefer compact symbolic forms. Since the six joints in \Cref{eq:stribeck_friction_model-1} share similar structures, we focus on symbolic identification for Joints 1 and 2.

\begin{figure}
	\begin{centering}
		\includegraphics[scale=0.40]{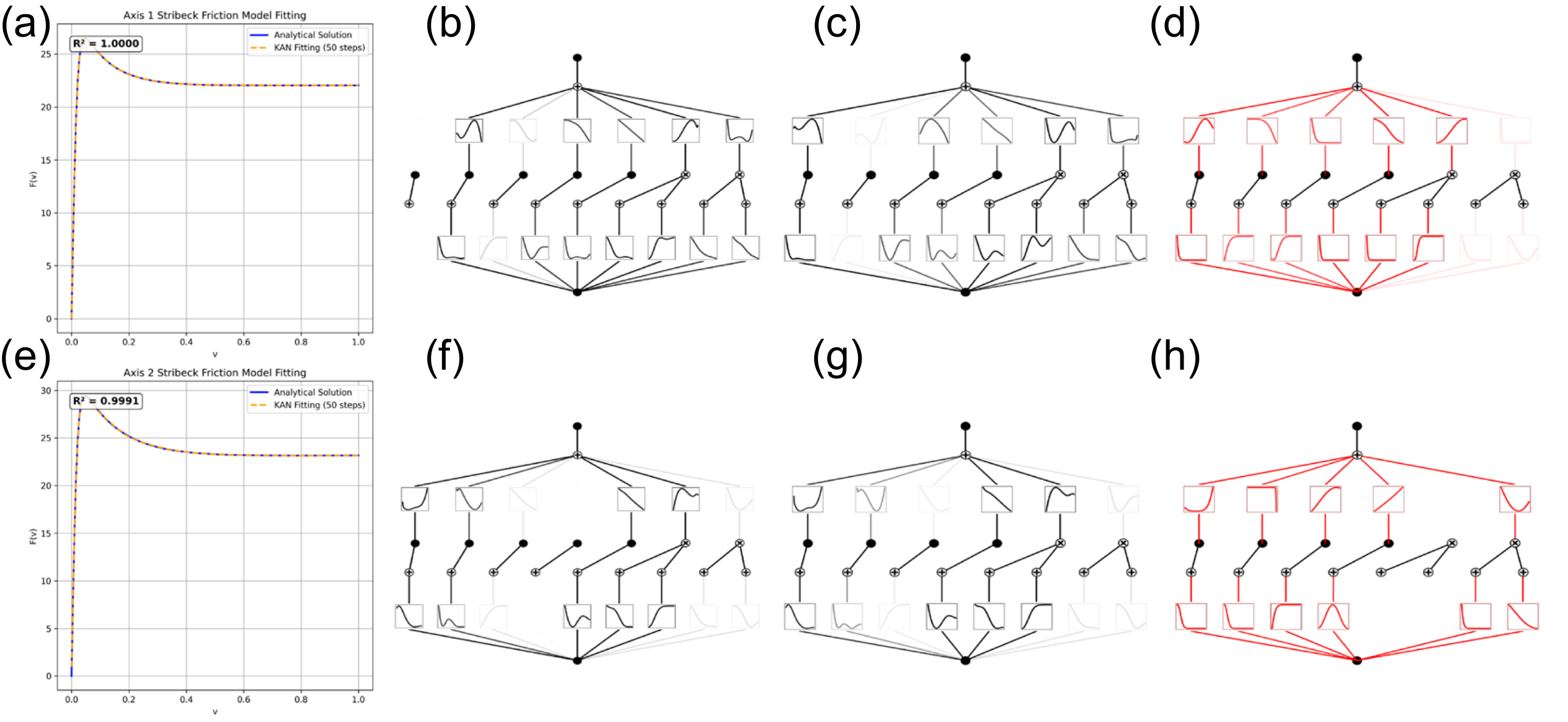}
		\par\end{centering}
	\caption{Symbolic identification of friction models for Joints 1 and 2. Rows 1 and 2 correspond to Joints 1 and 2, respectively. Column 1: comparison between the ground-truth solution and the KAN-derived symbolic solution. Column 2: KAN computational graph before pruning. Column 3: KAN graph after pruning. Column 4: KAN graph after automatic symbolic selection, where red nodes indicate fixed (selected) activations. In Columns 2–4, node/edge transparency encodes attribution scores (importance). \label{fig:KAN_computational_graph}}
\end{figure}

Because \Cref{eq:stribeck_friction_model-1} contains multiplicative interactions, a purely additive KAN can make symbolic identification cumbersome. We therefore adopt KAN with multiplicative nodes, with the initial architecture $[1,[5,2],1]$, where $[5,2]$ denotes five additive and two multiplicative nodes in the first hidden layer. After initializing KAN, we fit it to data, compute attribution scores, perform pruning (details in \Cref{subsec:KAN_inverse}), and refit. We then search a predefined function library to find expressions that best match the learned B-spline activations and refit once more using the symbolic forms. Each fitting stage uses 50 L-BFGS steps with learning rate $1.0$ (the learning rate can be tuned in practice). \Cref{fig:KAN_computational_graph} presents the symbolic results. The coefficient of determination is close to 1, indicating that KAN attains excellent symbolic identification even when the functional form is unknown.

\begin{figure}
	\begin{centering}
		\includegraphics[scale=0.38]{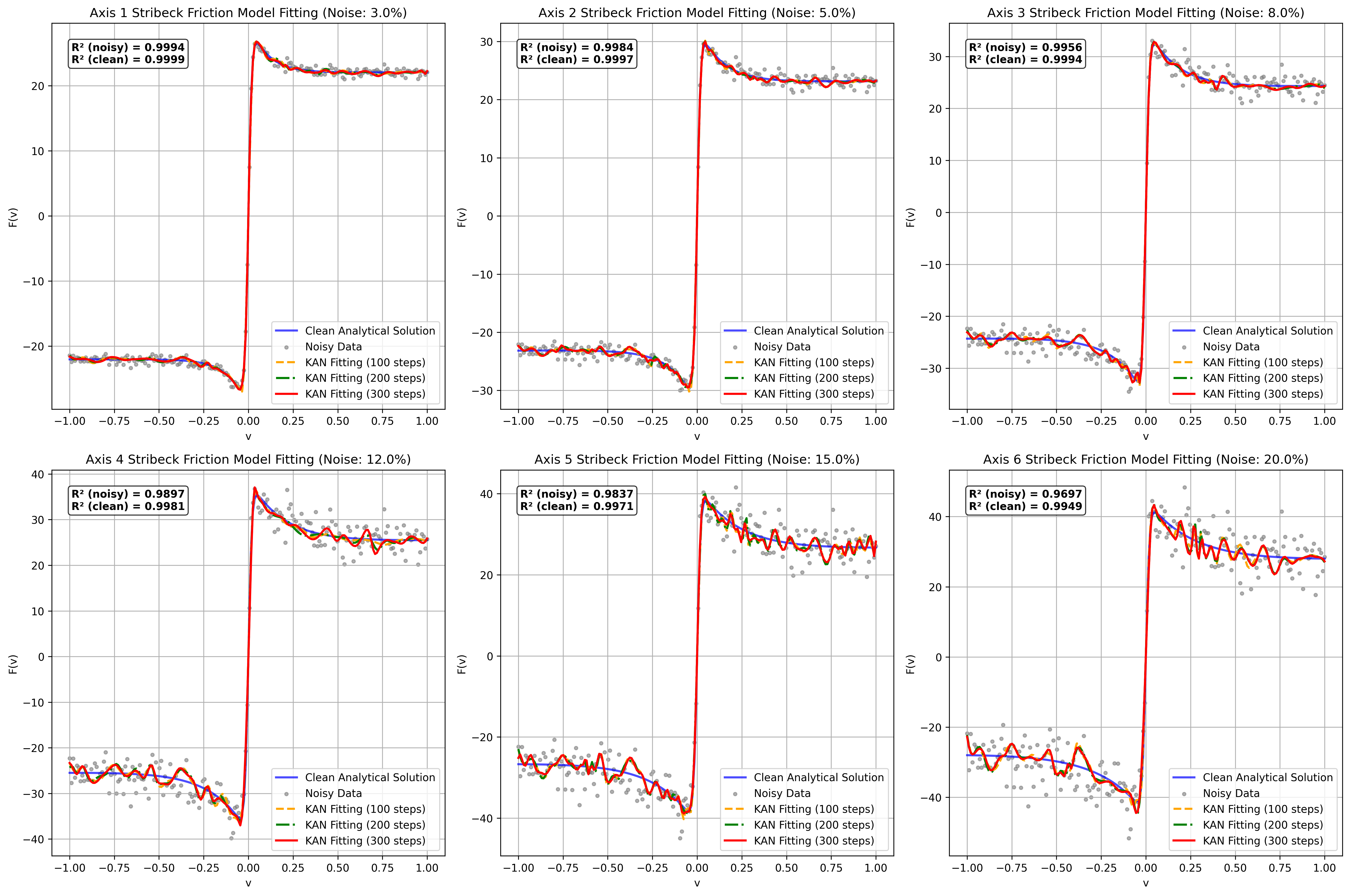}
		\par\end{centering}
	\caption{KAN fitting of friction models with unknown forms under noisy data. \label{fig:KAN_fitting_unknown_noisy}}
\end{figure}

\begin{figure}
	\begin{centering}
		\includegraphics[scale=0.45]{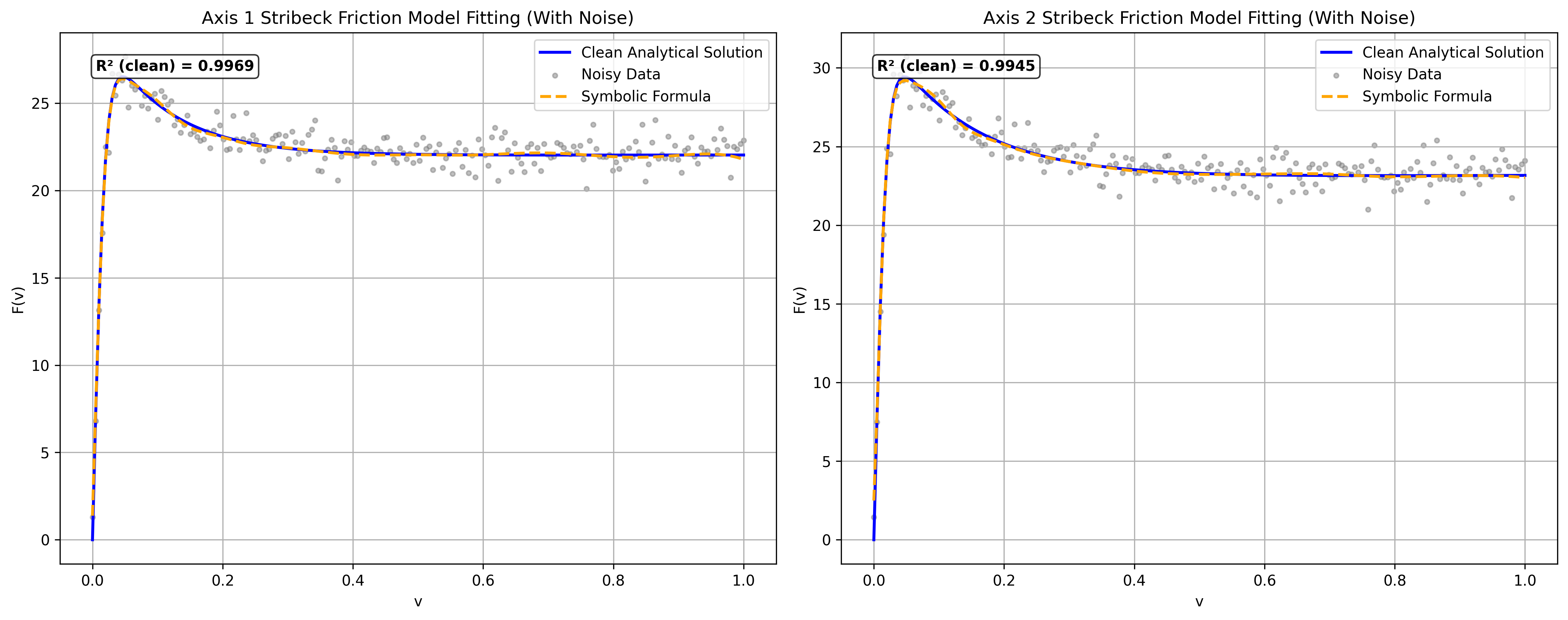}
		\par\end{centering}
	\caption{Symbolic identification by KAN under $5\%$ noise for unknown-form friction models. \label{fig:KAN_symbolic_noisy}}
\end{figure}

We next examine the model’s robustness to noise. Noise is applied as
\begin{equation}
	\mathrm{Std}=\frac{\lambda}{2}\,[\max(F)-\min(F)],
\end{equation}
with Gaussian perturbations $N(0,\mathrm{Std})$. The noise levels $\lambda$ for Joints 1–6 are set to $[3\%,\,5\%,\,8\%,\,12\%,\,15\%,\,20\%]$. \Cref{fig:KAN_fitting_unknown_noisy} shows fitting performance under noise, while \Cref{fig:KAN_symbolic_noisy} illustrates symbolic recovery with $5\%$ noise. The resulting symbolic expressions are summarized in \Cref{tab:auto_KAN}. The results indicate that KAN maintains strong robustness in symbolic identification in the presence of noise.

\begin{table}
	\caption{Fully automated symbolic regression results obtained by KAN. \label{tab:auto_KAN}}
	\centering{}%
	\begin{adjustbox}{max width=\textwidth}
		\begin{tabular}{ccc}
			\toprule 
			Axis & Noise & Friction model \tabularnewline
			\midrule
			\multirow{10}{*}{1} & Truth & $(22+8\exp[-|\tfrac{v}{0.1}|])\tanh(50v)+0.03v$ \tabularnewline
			\cmidrule{2-3}
			& \multirow{4}{*}{without Noisy} & \multicolumn{1}{c}{1.3419{*}sin(5.7236{*}(-0.9156 - 0.2478{*}exp(-49.0{*}(0.0783 - v){*}{*}2)){*}(-0.65
				+ 0.9254{*}exp(-38.44{*}(0.0071 - v){*}{*}2)) } \tabularnewline
			& & - 5.5839) - 2.0181{*}atan(1.7857{*}cos(7.5437{*}v - 5.7734) - 1.3159)
			+ 21.6801 - 14.9866{*}exp(-105.3328{*}(0.0722{*}tan(1.4{*}v - 7.788)
			+ 1){*}{*}2) \tabularnewline
			& & + 3.1463{*}exp(-88.36{*}(-(2.2205{*}sin(2.268{*}v + 9.5618) + 0.4448){*}(-1.1587{*}tanh(1.1991{*}v
			- 0.324) - 0.395) - 0.0387){*}{*}2) \tabularnewline
			& & + 0.1285/(-0.3133{*}sin(10.0{*}v + 0.6) - 1){*}{*}5 - 0.0036/(0.464
			+ exp(-20.3639{*}(0.3116 - v){*}{*}2)){*}{*}5 \tabularnewline
			\cmidrule{2-3}
			& \multirow{5}{*}{Noisy: $5\%$} & -64.3136{*}(-(0.2243 - 0.913{*}atan(4.9467{*}v - 0.2755)){*}(1.3497{*}tanh(9.4543{*}v
			- 1.1825) - 0.9366) - 0.1063){*}{*}5 \tabularnewline
			& & - 8.3805{*}cos(2.3492 + 0.132{*}exp(-4.7194{*}v)) + 6.6043 + 0.4933{*}exp(-19.5146{*}(-0.9964{*}tanh(5.2178{*}v
			- 2.0032) - 1){*}{*}2) \tabularnewline
			& & + 27.3305{*}exp(-1.0289{*}(-0.0037{*}sin(8.6814{*}v - 8.1879) - 1){*}{*}2)
			- 0.1784{*}exp(-26.0697{*}(0.9743 - cos(1.1717{*}v - 0.9035)){*}{*}2) \tabularnewline
			& & + 65.3711{*}exp(-5.4833{*}(-0.0286 - 1.3515{*}exp(-2.2982{*}(0.5761
			- v){*}{*}2)){*}(1.2045{*}sin(1.4438{*}v + 9.7541) - 0.2474)) \tabularnewline
			& & - 0.e-4/(-0.9077{*}sin(7.2203{*}v + 2.0928) - 1){*}{*}2 \tabularnewline
			\midrule
			\multirow{8}{*}{2} & Truth & $(23+10\exp[-|\tfrac{v}{0.13}|])\tanh(50v)+0.16v$ \tabularnewline
			\cmidrule{2-3}
			& \multirow{2}{*}{without Noisy} & 20.3664{*}(1 + 0.7159{*}exp(-12.8081{*}(-v - 0.2171){*}{*}2)){*}{*}5
			- 16.0797{*}Abs(0.0631 - 211.7302{*}exp(-71.3675{*}(-v - 0.2228){*}{*}2)) \tabularnewline
			& & + 3.8069 - 30.9582{*}exp(-107.3412{*}(-(2.9951 - 3.2095{*}exp(-0.1186{*}(1
			- 0.4232{*}v){*}{*}2)){*}(-1.6015{*}sin(5.2959{*}v - 8.5764) - 4.4884)
			- 0.3829){*}{*}2) \tabularnewline
			\cmidrule{2-3}
			& \multirow{5}{*}{Noisy: $5\%$} & -0.096{*}(-(1.2557{*}sin(4.8721{*}v + 0.0051) - 0.58){*}(-1.2702{*}tanh(2.479{*}v-
			0.8536) - 0.6114) + 0.3168){*}{*}2 \tabularnewline
			& & + 2.1757{*}sin(11.0535 - 3.8877{*}exp(-6.901{*}(-v - 0.2577){*}{*}2))
			+ 0.2632{*}Abs(1.4404{*}sin(7.5788{*}v - 6.1086) + 8.859) + 20.1143 \tabularnewline
			& & + 2.4195{*}exp(-714.8574{*}(0.9769{*}sin(4.7593{*}v - 1.8103) + 1){*}{*}2)
			- 2.5189{*}exp(-217.7511{*}(1 + 0.15{*}exp(-71.847{*}(-v - 0.2073){*}{*}2)){*}{*}2) \tabularnewline
			& & + 4.9522{*}exp(-0.4069{*}(-1 + 0.6342{*}exp(-51.6851{*}(0.0401 - v){*}{*}2)){*}{*}2)
			- 0.e-4/(-(-0.2497 - 2.2441{*}exp(-8.6454{*}(0.1346 -v){*}{*}2)) \tabularnewline
			& & {*}(0.0733 - 1.4005{*}exp(-7.4173{*}(0.8046 - v){*}{*}2)) - 0.1354){*}{*}2 \tabularnewline
			\bottomrule
		\end{tabular}
	\end{adjustbox}
\end{table}

\subsubsection{Performance on Real Data}

\begin{figure}
	\begin{centering}
		\includegraphics[scale=0.38]{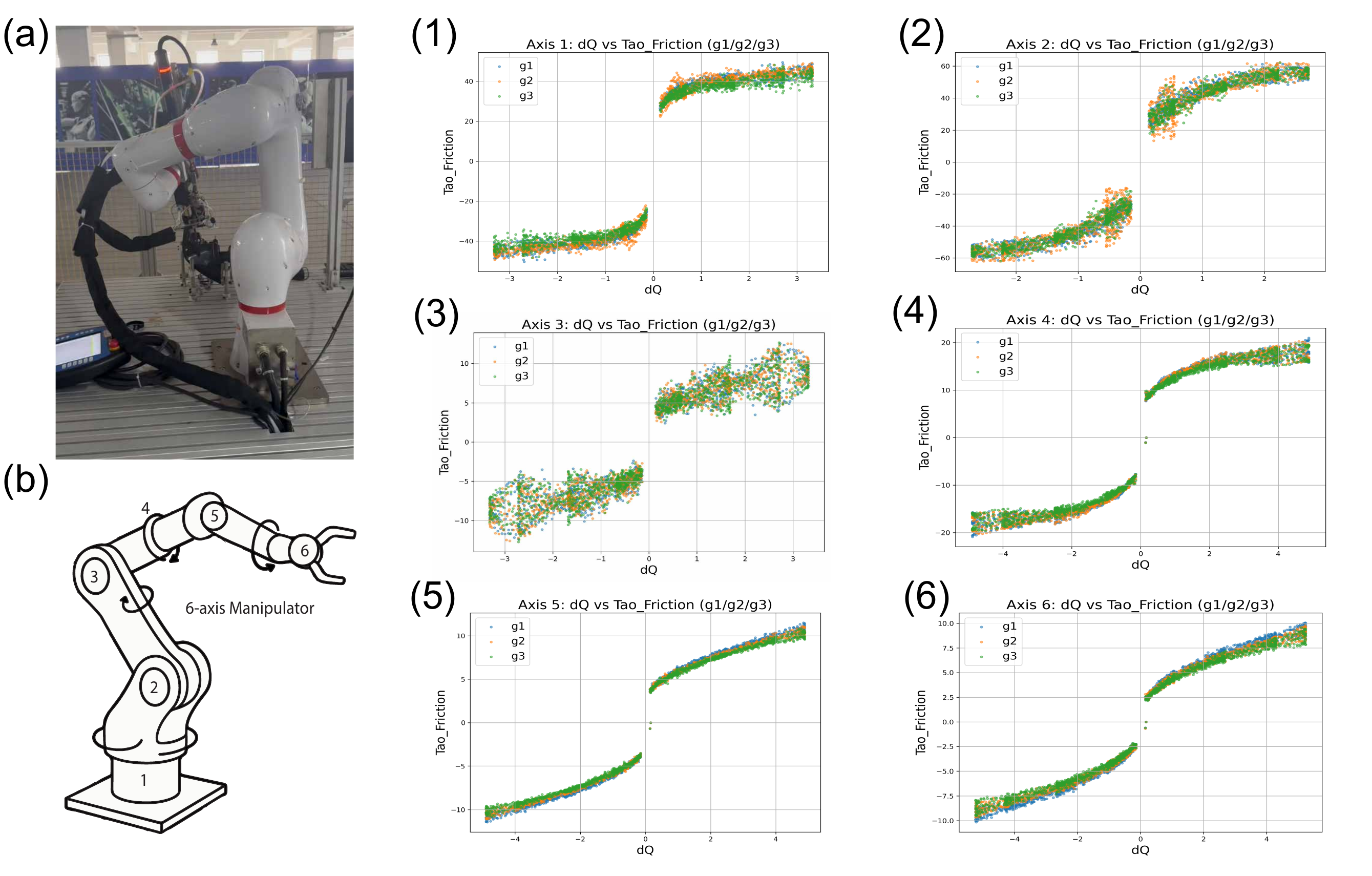}
		\par\end{centering}
	\caption{HSR-JR607 industrial manipulator with six axes. The horizontal axis is angular velocity, and the vertical axis is friction. g1, g2, and g3 denote different trajectories (configurations) for the same joint. \label{fig:HSR-JR607_experiment_data}}
\end{figure}

We next validate the feasibility of learning friction models from real measurements. The dataset is collected from an HSR-JR607 industrial manipulator with six joints (axes), as shown in \Cref{fig:HSR-JR607_experiment_data}. The friction data distributions differ across trajectories (denoted g1, g2, g3). In practical settings, it is common to learn a friction model on one trajectory and evaluate its generalization on others.

\begin{figure}
	\begin{centering}
		\includegraphics[scale=0.38]{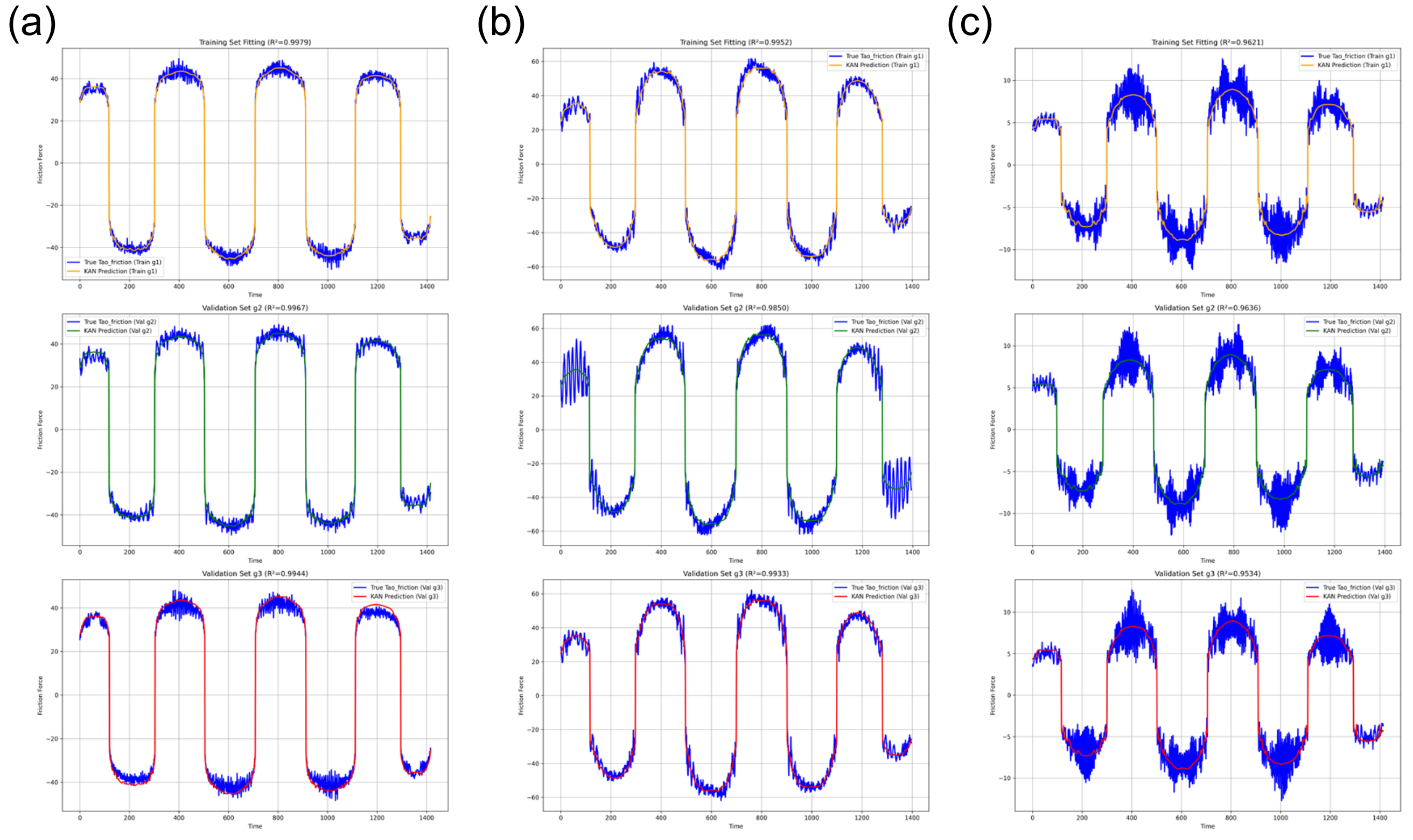}
		\par\end{centering}
	\caption{KAN learning on real data for Axis 1 (a), Axis 2 (b), and Axis 3 (c). Each axis has three trajectories (g1–g3). Training is performed on g1; g2 and g3 are used for testing. Row 1: training performance on g1. Rows 2–3: test performance on g2 and g3. \label{fig:KAN_to_real_data_friction_learning}}
\end{figure}

We investigate KAN-based learning on real data from Axes 1–3. Each axis contains three trajectories (g1–g3). We first fit the friction model on g1 and then test on g2 and g3. \Cref{fig:KAN_to_real_data_friction_learning} shows the learning results across different axes; all scenarios are well captured, with $R^{2}$ exceeding $0.95$. The KAN architecture is $[1,5,5,1]$ with 10 grid points per dimension and spline order 3. We train using L-BFGS for 30 steps. Each trajectory contains approximately 1,400 samples (collected at a fixed interval of 4 ms).

\begin{figure}
	\begin{centering}
		\includegraphics[scale=0.45]{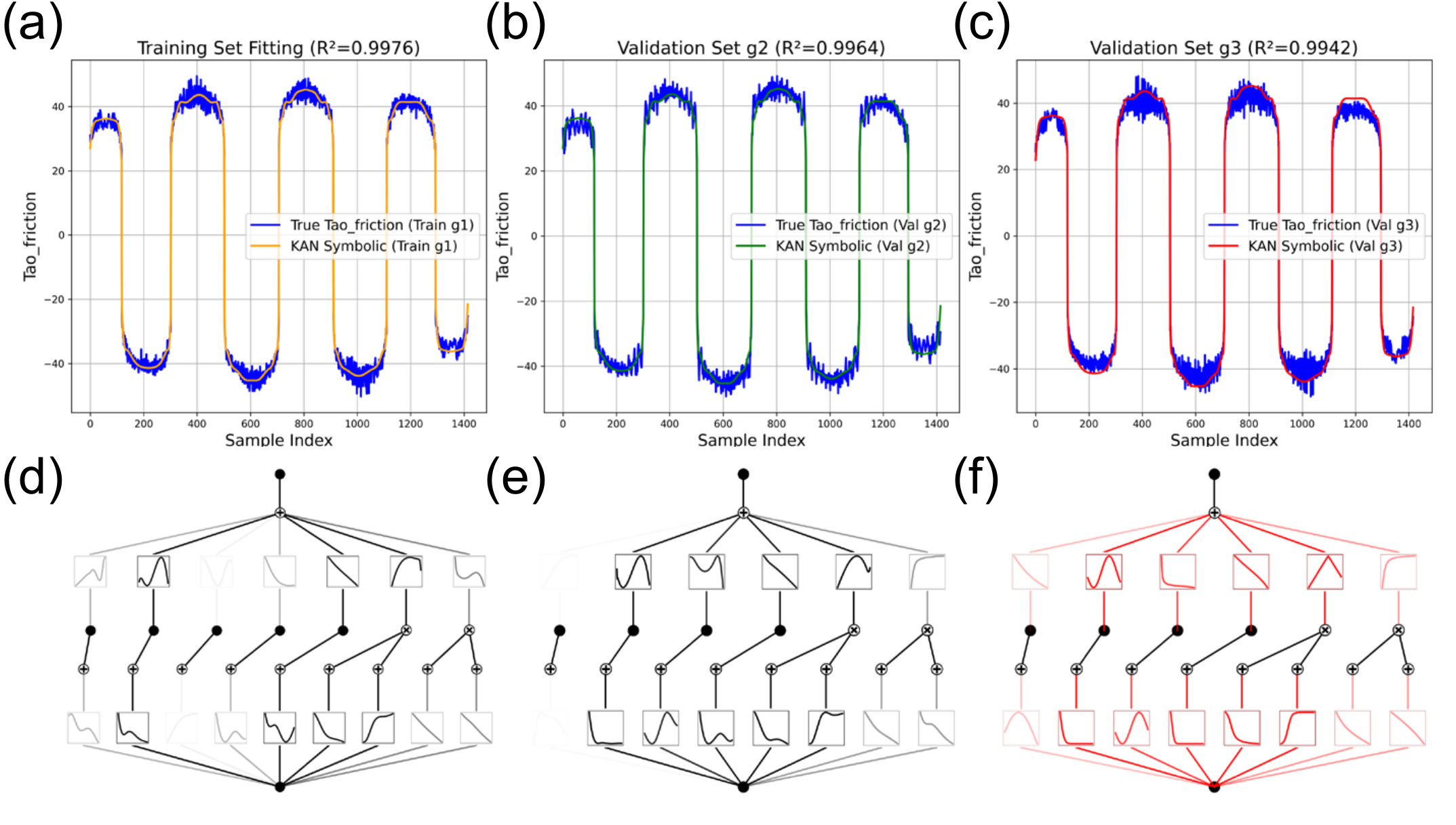}
		\par\end{centering}
	\caption{Symbolic regression on real data for Axis 1: (a) training on g1 and performance on the g1 training set; (b,c) test performance on g2 and g3; (d) KAN computational graph fitted on g1; (e) pruned KAN graph; (f) KAN graph after symbolic selection. \label{fig:KAN_to_real_data_friction_learning_symbolic}}
\end{figure}

We further evaluate symbolic regression on real data. The initial KAN with multiplicative nodes is configured as $[1,[5,2],1]$ (five additive and two multiplicative nodes in the first hidden layer), using 3 grid points per dimension and spline order 3. Training uses L-BFGS for 50 steps. \Cref{fig:KAN_to_real_data_friction_learning_symbolic} reports the results for Axis 1: KAN achieves strong symbolic recovery on real measurements. The resulting symbolic expressions are summarized in \Cref{tab:auto_KAN_real_data}.

\begin{table}
	\caption{Fully automated symbolic regression results by KAN on real data. \label{tab:auto_KAN_real_data}}
	\centering{}%
	\begin{adjustbox}{max width=\textwidth}
		\begin{tabular}{cc}
			\toprule 
			Axis & Friction model \tabularnewline
			\midrule 
			\multirow{7}{*}{1} & 3747.49598884058{*}sqrt(1 - 0.0210462241624075{*}atan(1.8004777431488{*}x\_1
			+ 0.0344132147729397))\tabularnewline
			& + 6.13821649551392{*}cos(0.0399466380476952{*}(13.353551864624{*}cos(0.690846920013428{*}x\_1
			- 1.47569036483765) \tabularnewline
			& + 1.44823837280273){*}(-8.60865211486816{*}tanh(3.43028998374939{*}x\_1
			+ 0.985089778900146) - 1.84565925598145) + 5.926833152771) \tabularnewline
			& - 56.8303375244141{*}cos(0.457886635704487{*}sin(0.436841130256653{*}x\_1
			- 3.39227294921875) + 8.22667917604575) \tabularnewline
			& + 96.1950988769531{*}tanh(0.0605322915946442{*}tan(0.26903161406517{*}x\_1
			+ 0.431818008422852) - 0.143116033216823) \tabularnewline
			& + 81.6320877075195{*}tanh(1.33775630231523{*}tanh(2.48217296600342{*}x\_1
			+ 0.0918278321623802) - 0.192893696946026) - 3912.35780066252 \tabularnewline
			& + 236.195907592773{*}exp(-0.406416481602622{*}(0.294385951896575{*}sin(0.449970006942749{*}x\_1
			+ 3.08874845504761) + 1){*}{*}2)\tabularnewline
			\midrule 
			\multirow{9}{*}{single to multi} & 1.1329478945244{*}(-0.096910859162038{*}(-3.25447535514832 + 7.77069711685181{*}exp(-5.2471151520927{*}(-0.679981727188676{*}x\_1
			- 1){*}{*}2))\tabularnewline
			& {*}(0.33010059595108 - 4.01152753829956{*}cos(1.52239143848419{*}x\_1
			- 4.44621515274048)) - 1){*}{*}2 \tabularnewline
			& + 4005.11346983425{*}exp(0.0640550957834551{*}exp(-0.389798200002819{*}(-0.768977972262003{*}x\_1
			- 1){*}{*}2)) \tabularnewline
			& - 124.540855407715{*}cos(6.45286869389031 - 7.97604270966824/sqrt(1
			- 0.0866478331886679{*}x\_1))\tabularnewline
			& + 0.80953973531723{*}tanh(9.99937343597412{*}(0.739961743354797{*}cos(0.786135494709015{*}x\_1
			+ 7.79666090011597) \tabularnewline
			& + 0.463700741529465){*}(0.15334589779377{*}Abs(7.54662752151489{*}x\_1
			+ 8.56246376037598) - 0.919431447982788) \tabularnewline
			& + 1.00389671325684) + 170.220413208008{*}tanh(0.966172500974238{*}cos(1.68652188777924{*}x\_1
			- 1.67582201957703) + 0.174910167612076) \tabularnewline
			& - 141.846252441406{*}atan(0.318889115657537{*}sin(1.90128135681152{*}x\_1
			+ 3.04936456680298) + 0.276258981896176) - 7149.87543821335\tabularnewline
			& + 6161.443359375{*}exp(-0.720039308267579{*}(0.0198799710312073{*}sin(2.04783225059509{*}x\_1
			+ 6.27415037155151) + 1){*}{*}2)\tabularnewline
			\bottomrule
		\end{tabular}
	\end{adjustbox}
\end{table}

\begin{figure}
	\begin{centering}
		\includegraphics[scale=0.35]{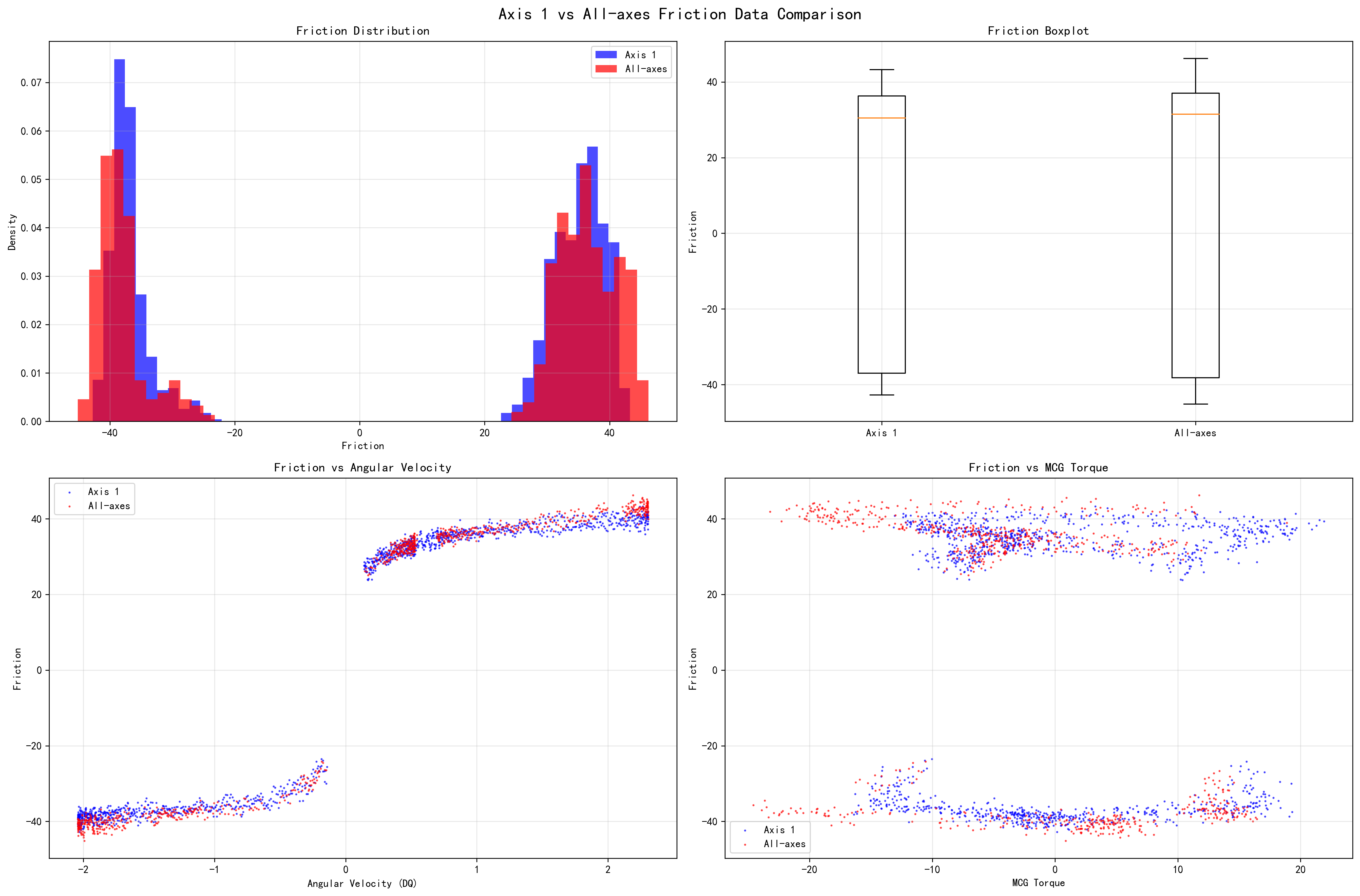}
		\par\end{centering}
	\caption{Single- vs multi-axis data distributions. Top row: friction data distributions under single- and multi-axis motions. Bottom row (left to right): scatter of angular velocity vs friction; scatter of joint torque $\tau_{\mathrm{mcg}}$ vs friction. \label{fig:single_multi}}
\end{figure}

In real applications, one often learns a friction model from \emph{single-axis} data and then generalizes it to \emph{multi-axis} motions. Here, “single-axis’’ means only one joint moves while others are fixed, whereas “multi-axis’’ means all joints move simultaneously. The data again come from the HSR-JR607 manipulator. \Cref{fig:single_multi} compares the distributions of single- and multi-axis measurements. The joint torque $\tau_{\mathrm{mcg}}$ denotes the torque induced by the mass–inertia (M), Coriolis (C), and gravity (G) terms in robot dynamics. The correlation coefficients between friction and angular velocity are $0.9196$ (single-axis) and $0.9306$ (multi-axis). The correlations between friction and $\tau_{\mathrm{mcg}}$ are $0.0219$ (single-axis) and $-0.3749$ (multi-axis). Thus, friction correlates much more strongly with angular velocity than with $\tau_{\mathrm{mcg}}$, and we choose angular velocity as the sole input variable. That said, in other scenarios $\tau_{\mathrm{mcg}}$ may play a critical role; a preliminary correlation analysis is recommended to determine whether $\tau_{\mathrm{mcg}}$ should be included as an input.

\begin{figure}
	\begin{centering}
		\includegraphics[scale=0.36]{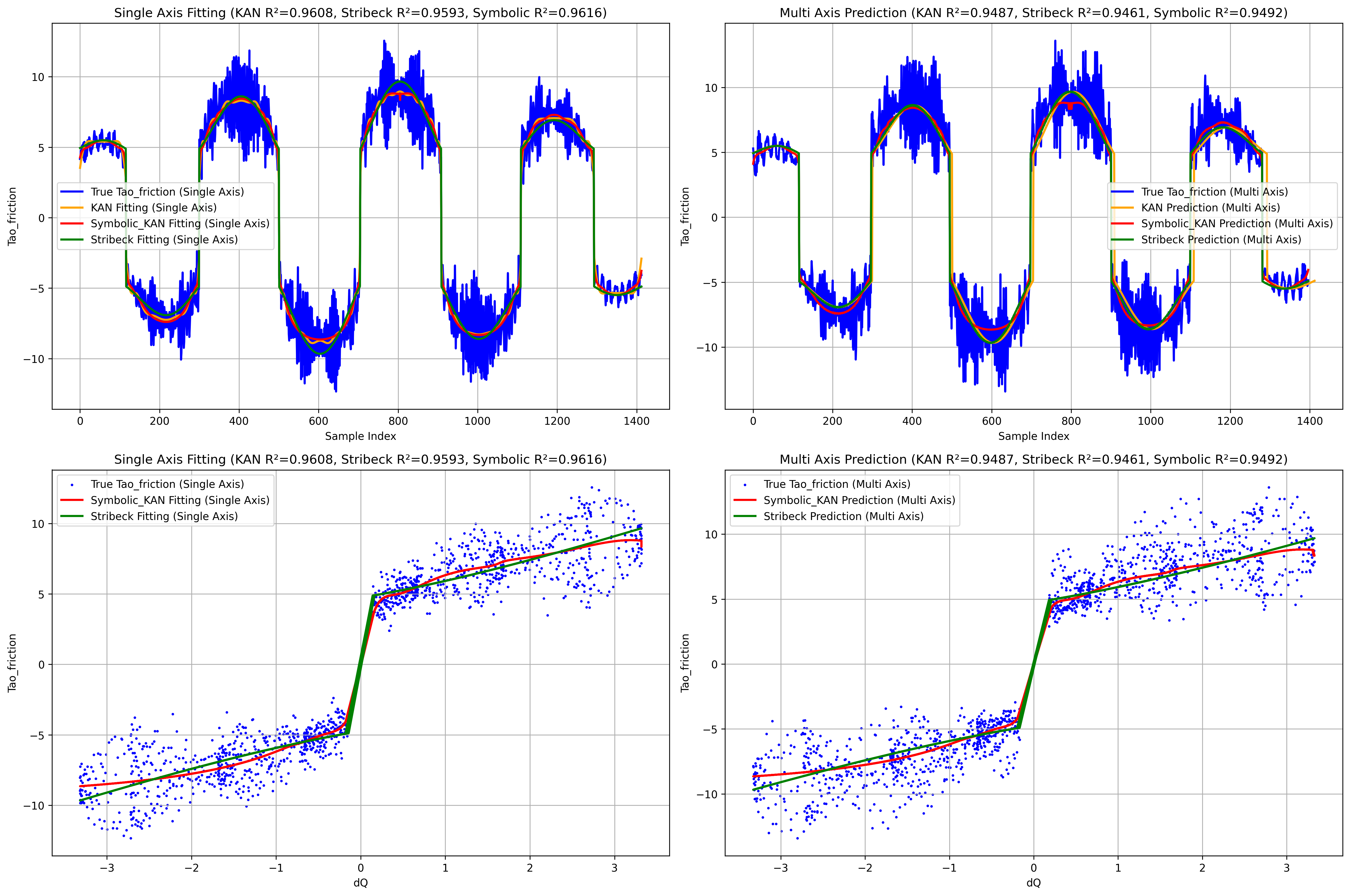}
		\par\end{centering}
	\caption{Single-axis (Axis 3) training and multi-axis generalization. Column 1: learning on single-axis data. Column 2: prediction on multi-axis data. “KAN fitting’’ denotes KAN fit on single-axis data; “KAN prediction’’ denotes KAN prediction on multi-axis data. “Symbolic\_KAN’’ is the KAN prediction after symbolic regression; “Stribeck model’’ is the fit using the known functional form. Row 1 uses time as the horizontal axis; Row 2 uses angular velocity. \label{fig:single2multi_symbolic}}
\end{figure}

\Cref{fig:single2multi_symbolic} demonstrates generalization from a model learned on single-axis data (Axis 3) to multi-axis motions. The symbolic KAN behaves consistently with the known-form Stribeck model in \Cref{subsec:known_function}. Notably, in this subsection the functional form is \emph{unknown}; KAN searches a large hypothesis space and then returns compact symbolic expressions (see \Cref{tab:auto_KAN_real_data}).

\section{Conclusion} \label{sec:conclusion}

This paper presents a physics-inspired machine learning framework for static friction modeling and symbolic expression in robotic systems, built upon the Kolmogorov--Arnold Network (KAN) architecture. 
The proposed approach combines the high predictive accuracy of data-driven learning with the physical interpretability of symbolic regression, making it particularly suitable for modeling joint friction in industrial manipulators. 

Through extensive experiments, we demonstrate that KAN can automatically generate friction expressions that closely align with physical laws, even when the underlying functional form is completely unknown. 
By incorporating network pruning and symbolic regression, the resulting models are both compact and physically meaningful. 
Moreover, they exhibit strong generalization capability: a model trained on one trajectory generalizes effectively to others, and a single-axis model can be extended to multi-axis motions. 
In real robotic experiments, the proposed approach consistently achieved coefficients of determination ($R^2$) exceeding 0.95, confirming its robustness and practical viability.

Overall, this study introduces a new paradigm for interpretable, accurate, and deployable friction modeling in robotics. 
Future work will focus on extending the framework to dynamic friction modeling, incorporating additional physical variables (e.g., joint torque, temperature) for comprehensive joint behavior characterization, and embedding the symbolic friction models into real-time control loops to enable online friction compensation and adaptive control.

\section*{Declaration of competing interest}
The authors declare that they have no known competing financial interests or personal relationships that could
have appeared to influence the work reported in this paper.

\section*{Acknowledgement}
The study was supported by the Key Project of the National Natural Science Foundation of China (12332005).

\bibliographystyle{elsarticle-num}
\addcontentsline{toc}{section}{\refname}\bibliography{friction.bib}

@article{ill_gradient,
   author = {Wang, Sifan and Teng, Yujun and Perdikaris, Paris %J SIAM Journal on Scientific Computing},
	title = {Understanding and mitigating gradient flow pathologies in physics-informed neural networks},
	journal = {SIAM Journal on Scientific Computing},
	volume = {43},
	number = {5},
	pages = {A3055-A3081},
	ISSN = {1064-8275},
	year = {2021},
	type = {Journal Article}
}

@article{ALEXNET,
   author = {Krizhevsky, Alex and Sutskever, Ilya and Hinton, Geoffrey E},
   title = {Imagenet classification with deep convolutional neural networks},
   journal = {Advances in neural information processing systems},
   volume = {25},
   pages = {1097--1105},
   year = {2012},
   type = {Journal Article}
}

@article{speech_recognition,
   author = {Graves, Alex and Fern{\'a}ndez, Santiago and Gomez, Faustino and Schmidhuber, J{\"u}rgen},
   title = {Connectionist temporal classification: labelling unsegmented sequence data with recurrent neural networks},
   pages = {369--376},
   year = {2006},
   type = {Journal Article}
}

@article{alphago,
   author = {Silver, David and Huang, Aja and Maddison, Chris J and Guez, Arthur and Sifre, Laurent and Van Den Driessche, George and Schrittwieser, Julian and Antonoglou, Ioannis and Panneershelvam, Veda and Lanctot, Marc and others},
   title = {Mastering the game of Go with deep neural networks and tree search},
   journal = {Nature},
   volume = {529},
   number = {7587},
   pages = {484-489},
   note = {OA status: bronze},
   ISSN = {0028-0836},
   DOI = {10.1038/nature16961},
   url = {https://www.nature.com/articles/nature16961.pdf},
   year = {2016},
   type = {Journal Article}
}

@article{star_game,
   author = {Vinyals, Oriol and Babuschkin, Igor and Czarnecki, Wojciech M and Mathieu, Micha{\"e}l and Dudzik, Andrew and Chung, Junyoung and Choi, David H and Powell, Richard and Ewalds, Timo and Georgiev, Petko and others},
   title = {Grandmaster level in StarCraft II using multi-agent reinforcement learning},
   journal = {Nature},
   volume = {575},
   number = {7782},
   pages = {350-354},
   ISSN = {0028-0836},
   DOI = {10.1038/s41586-019-1724-z},
   year = {2019},
   type = {Journal Article}
}

@article{alphafold,
   author = {Senior, Andrew W and Evans, Richard and Jumper, John and Kirkpatrick, James and Sifre, Laurent and Green, Tim and Qin, Chongli and {\v{Z}}{\'\i}dek, Augustin and Nelson, Alexander WR and Bridgland, Alex and others},
   title = {Improved protein structure prediction using potentials from deep learning},
   journal = {Nature},
   volume = {577},
   number = {7792},
   pages = {706-710},
   ISSN = {0028-0836},
   DOI = {10.1038/s41586-019-1923-7},
   year = {2020},
   type = {Journal Article}
}

@article{li2019predicting,
  title={Predicting the effective mechanical property of heterogeneous materials by image based modeling and deep learning},
  author={Li, Xiang and Liu, Zhanli and Cui, Shaoqing and Luo, Chengcheng and Li, Chenfeng and Zhuang, Zhuo},
  journal={Computer Methods in Applied Mechanics and Engineering},
  volume={347},
  pages={735--753},
  year={2019},
  publisher={Elsevier}
}

@inproceedings{he2016deep,
  title={Deep residual learning for image recognition},
  author={He, Kaiming and Zhang, Xiangyu and Ren, Shaoqing and Sun, Jian},
  booktitle={Proceedings of the IEEE conference on computer vision and pattern recognition},
  pages={770--778},
  year={2016}
}

@article{brown2020language,
  title={Language models are few-shot learners},
  author={Brown, Tom and Mann, Benjamin and Ryder, Nick and Subbiah, Melanie and Kaplan, Jared D and Dhariwal, Prafulla and Neelakantan, Arvind and Shyam, Pranav and Sastry, Girish and Askell, Amanda and others},
  journal={Advances in neural information processing systems},
  volume={33},
  pages={1877--1901},
  year={2020}
}

@article{liu2024kan,
  title={Kan: Kolmogorov-arnold networks},
  author={Liu, Ziming and Wang, Yixuan and Vaidya, Sachin and Ruehle, Fabian and Halverson, James and Soljacc, Marin and Hou, Thomas Y and Tegmark, Max},
  journal={arXiv preprint arXiv:2404.19756},
  year={2024}
}

@article{liu2024kan2,
  title={Kan 2.0: Kolmogorov-arnold networks meet science},
  author={Liu, Ziming and Ma, Pingchuan and Wang, Yixuan and Matusik, Wojciech and Tegmark, Max},
  journal={arXiv preprint arXiv:2408.10205},
  year={2024}
}

@article{lu2006stribeck,
  title={The Stribeck curve: experimental results and theoretical prediction},
  author={Lu, Xiaobin and Khonsari, MM and Gelinck, ERM},
  journal={Journal of tribology},
  volume={128},
  number={4},
  pages={789--794},
  year={2006}
}

@article{he2017experimental,
  title={Experimental and numerical investigations of the stribeck curves for lubricated counterformal contacts},
  author={He, Tao and Zhu, Dong and Wang, Jiaxu and Jane Wang, Q},
  journal={Journal of Tribology},
  volume={139},
  number={2},
  pages={021505},
  year={2017},
  publisher={American Society of Mechanical Engineers}
}

@article{bo1982friction,
  title={The friction-speed relation and its influence on the critical velocity of stick-slip motion},
  author={Bo, Li Chun and Pavelescu, D},
  journal={Wear},
  volume={82},
  number={3},
  pages={277--289},
  year={1982},
  publisher={Elsevier}
}

@article{marques2016survey,
  title={A survey and comparison of several friction force models for dynamic analysis of multibody mechanical systems},
  author={Marques, Filipe and Flores, Paulo and Pimenta Claro, JC and Lankarani, Hamid M},
  journal={Nonlinear Dynamics},
  volume={86},
  pages={1407--1443},
  year={2016},
  publisher={Springer}
}

@article{hess1990friction,
  title={Friction at a lubricated line contact operating at oscillating sliding velocities},
  author={Hess, DP and Soom, Andres},
  year={1990}
}

@article{coulomb1781theorie,
  title={Th{\'e}orie des machines simples, en ayant {\'e}gard au frottement de leurs parties; et a la roideur des cordages.},
  author={Coulomb, Charles Augustin},
  year={1781}
}

@article{bengisu1994stability,
  title={Stability of friction-induced vibrations in multi-degree-of-freedom systems},
  author={Bengisu, MT and Akay, A},
  journal={Journal of Sound and Vibration},
  volume={171},
  number={4},
  pages={557--570},
  year={1994},
  publisher={Elsevier}
}

@article{awrejcewicz2008404,
  title={404. A novel dry friction modeling and its impact on differential equations computation and Lyapunov exponents estimation.},
  author={Awrejcewicz, Jan and Grzelczyk, D and Pyryev, Yu},
  journal={Journal of Vibroengineering},
  volume={10},
  number={4},
  year={2008}
}

@inproceedings{armstrong1992frictional,
  title={Frictional lag and stick-slip},
  author={Armstrong-Helouvry, Brian},
  booktitle={Proceedings 1992 IEEE International Conference on Robotics and Automation},
  pages={1448--1449},
  year={1992},
  organization={IEEE Computer Society}
}

@inproceedings{lampaert2003generalized,
  title={A generalized Maxwell-slip friction model appropriate for control purposes},
  author={Lampaert, Vincent and Al-Bender, Farid and Swevers, Jan},
  booktitle={2003 IEEE International Workshop on Workload Characterization (IEEE Cat. No. 03EX775)},
  volume={4},
  pages={1170--1177},
  year={2003},
  organization={IEEE}
}

@article{piatkowski2014dahl,
  title={Dahl and LuGre dynamic friction models—The analysis of selected properties},
  author={Piatkowski, Tomasz},
  journal={Mechanism and Machine Theory},
  volume={73},
  pages={91--100},
  year={2014},
  publisher={Elsevier}
}

\end{document}